\documentclass[preprint,12pt,authoryear]{elsarticle}

\usepackage{amssymb}
\usepackage{gensymb}
\usepackage{graphicx}
\usepackage{subcaption}
\usepackage{comment}
\usepackage{hyperref}
\usepackage{multirow}
\usepackage{rotating}
\usepackage{forest}
\usepackage{listings}
\usepackage{xcolor}
\definecolor{myblue}{RGB}{0,0,0}
\usepackage{enumitem}
\usepackage{etoolbox}
\patchcmd{\pprintMaketitle}
 {\ifvoid\absbox\else\unvbox\absbox\par\vskip10pt\fi}
 {\ifvoid\absbox\else\clearpage\unvbox\absbox\par\vskip30pt\fi}
 {}{}
\patchcmd{\pprintMaketitle}
 {\hrule\vskip12pt}
 {}
 {}{}
\patchcmd{\pprintMaketitle}
 {\hrule\vskip12pt}
 {}
 {}{}
\appto{\pprintMaketitle}{}
\lstdefinestyle{hierarchical}{
    basicstyle=\footnotesize\ttfamily,
    numbers=none,
    frame=none,
    breaklines=true,
    postbreak=\mbox{\textcolor{gray}{$\hookrightarrow$}\space},
    showstringspaces=false,
    keywordstyle=\bfseries\color{blue},
    keywords={If},
    emphstyle=\bfseries\color{purple},
    emph={[2]Choice:},
    commentstyle=\color{gray},
    emph={[3]Boolean},
    emphstyle={[3]\color{olive}},
    emph={[4]Choice:},
    emphstyle={[4]\color{magenta}},
    emph={[5]Parent condition:},
    emphstyle={[5]\color{teal}},
    tabsize=2,
    literate=*{--}{--}1
}

\lstdefinestyle{mystyle}{
    language=Python,
    style=hierarchical
}

\journal{Expert Systems with Applications}

\begin{document}

\begin{frontmatter}



\title{MoistNet: Machine Vision-based Deep Learning Models for Wood Chip Moisture Content Measurement}


\author[inst1]{Abdur Rahman}
\ead{ar2806@msstate.edu}
\affiliation[inst1]{organization={Department of Industrial and Systems Engineering},
            addressline={Mississippi State University}, 
            city={Mississippi State},
            postcode={MS 39762}, 
            country={USA}}
\author[inst2]{Jason Street}
\ead{jts118@msstate.edu}
\author[inst3]{James Wooten}
\ead{james.wooten@msstate.edu}
\author[inst1]{Mohammad Marufuzzaman}
\ead{maruf@ise.msstate.edu}
\author[inst4]{Veera G. Gude}
\ead{vgude@pnw.edu}
\author[inst5]{Randy Buchanan}
\ead{Randy.K.Buchanan@erdc.dren.mil}
\author[inst1]{Haifeng Wang\corref{cor1}}
\cortext[cor1]{Corresponding author: Haifeng Wang (wang@ise.msstate.edu)}
\ead{wang@ise.msstate.edu}
\affiliation[inst2]{organization={Department of Sustainable Bioproducts},
            addressline={Mississippi State University}, 
            city={Mississippi State},
            postcode={MS 39762}, 
            country={USA}}
\affiliation[inst3]{organization={Department of Agricultural and Biological Engineering},
            addressline={Mississippi State University}, 
            city={Mississippi State},
            postcode={MS 39762}, 
            country={USA}}
\affiliation[inst4]{organization={Purdue University Northwest Water Institute (PWI)},
            addressline={Purdue University Northwest}, 
            city={Hammond},
            postcode={IN 46323}, 
            country={USA}}

\affiliation[inst5]{organization={US Army Engineer Research and Development Center},
           addressline={3909 Halls Ferry Road},
            city={Vicksburg},
            postcode={39180}, 
            state={MS},
            country={USA}}

\begin{abstract}

Quick and reliable measurement of wood chip moisture content is an everlasting problem for numerous forest-reliant industries such as biofuel, pulp and paper, and bio-refineries. Moisture content is a critical attribute of wood chips due to its direct relationship with the final product quality. Conventional techniques for determining moisture content, such as oven-drying, possess some drawbacks in terms of their time-consuming nature, potential sample damage, and lack of real-time feasibility. Furthermore, alternative techniques, including NIR spectroscopy, electrical capacitance, X-rays, and microwaves, have demonstrated potential; nevertheless, they are still constrained by issues related to portability, precision, and the expense of the required equipment. {\color{myblue}Hence, there is a need for a moisture content determination method that is instant, portable, non-destructive, inexpensive, and precise.} This study explores the use of deep learning and machine vision to predict moisture content {\color{myblue}classes} from RGB images of wood chips. A large-scale image dataset comprising 1,600 RGB images of wood chips has been collected and annotated with ground truth labels, utilizing the results of the oven-drying technique. Two high-performing neural networks, MoistNetLite and MoistNetMax, have been developed leveraging Neural Architecture Search (NAS) and hyperparameter optimization. The developed models are evaluated and compared with state-of-the-art deep learning models. Results demonstrate that MoistNetLite achieves 87\% accuracy with minimal computational overhead, while MoistNetMax exhibits exceptional precision with a 91\% accuracy in wood chip moisture content class prediction. With improved accuracy (9.6\% improvement in accuracy by MoistNetMax compared to the best baseline model ResNet152V2) and faster prediction speed (MoistNetLite being twice as fast as MobileNet), our proposed MoistNet models hold great promise for the wood chip processing industry to be efficiently deployed on portable devices, such as smartphones. 
\end{abstract}








\begin{keyword}
Wood chip \sep moisture content \sep deep learning \sep machine vision \sep neural architecture search \sep hyperparameter optimization
\end{keyword}

\end{frontmatter}






\section{Introduction}
\label{sec:introduction}
Wood chips are crucial raw materials for various industries, including biofuel, pulp and paper, and bio-refineries. The moisture content (MC) of wood chips holds significant importance in all these industries. In biofuel production, the net calorific value of wood pellets, which determines the overall energy output, is influenced by the MC \citep{lev2021electrical}. Typically, as the MC increases, the net calorific value decreases, necessitating additional energy for moisture evaporation in wood chip processing \citep{daassi2018moisture}. Similarly, in the pulp and paper sector, the MC affects the concentration of chemicals utilized in lignin digestion, thereby impacting manufacturing process control. Hence, optimizing the manufacturing operations relies on precise and efficient MC measurement \citep{nystrom2004methods}. 
Therefore, classifying the delivered raw wood chip containers based on their MC is a crucial concern for these industries \citep{daassi2018moisture}.

Efficient MC determination remains a challenging task both in small- and large-scale manufacturing plants. The traditional oven-drying method, also known as the direct method, is still a widely recognized way to determine MC. Even though the direct method is one of the most precise methods, it has its own limitations. {\color{myblue}For instance, the direct method is destructive (samples are not reusable), slow, labor-intensive, and most importantly, it is not applicable in real-time \citep{lev2021electrical}.} Hence, a fast, non-destructive, easy-to-implement, and real-time applicable method is warranted to address these shortcomings.

Several indirect methods are being proposed to measure MC in this regard, including methods based on {\color{myblue}NIR spectroscopy \citep{nascimbem2013determination, liang2019determination, amaral2020estimation, toscano2022performance, yan2024moisture}, electrical capacitance \citep{kandala2016capacitance, lev2021electrical, fridh2018precision, jensen2006moisture, pan2016predicting, de2023dielectric}}, microwaves \citep{d2010simple, cazzorla2012woodchip, ottosson2018uwb}, X-rays \citep{kullenberg2010dual, hultnas2012determination, jain2017dual}, nuclear magnetic resonance (NMR) \citep{fridh2014accurate}, Wi-Fi \citep{suthar2021multiclass}, and images \citep{plankenbuhler2020image, rahmaninterpretable}. 

{\color{myblue}\citet{nascimbem2013determination} investigated quality parameter determination in moist wood chips using NIR spectroscopy and chemometrics, achieving a classification error of less than 6\% for MC determination with partial least squares-discriminant analysis (PLS-DA). Additionally, they developed reliable calibration models for quality parameters using least squares support vector machines (LS-SVM), showcasing the potential of NIR spectroscopy for wood chip quality control.} \citet{liang2019determination} determined the MC and basic density of poplar wood chips in various moisture conditions using NIR spectroscopy. \citet{amaral2020estimation} performed a similar NIR spectroscopy-based analysis on \textit{Eucalyptus} wood chips. \citet{toscano2022performance}, on the other hand, analyzed the performance of a portable NIR spectrometer to determine the MC of wood chips. {\color{myblue}While NIR spectroscopy has been recognized as a promising method for MC prediction, it requires a special scanning device (NIR spectrometer). Additionally, its applicability is limited to the wood chip surface and is affected by the size distribution and geometry of the wood chips \citep{liang2019determination, amaral2020estimation, toscano2022performance}}.

The dielectric properties of wood chips help develop capacitance-based methods to determine the MC. Using this theory, \citet{kandala2016capacitance} proposed an MC prediction method for hardwood chips. The authors observed better accuracy for samples that have MC less than 25\% \citep{kandala2016capacitance}. \citet{lev2021electrical} utilized an LCR (inductance (L), capacitance (C), and resistance (R)) meter to develop another capacitance-based method to predict the MC and porosity of wood chips. {\color{myblue}They developed linear models using backward stepwise linear regression with high accuracy ($R^2$ of 0.9–0.99). \citet{fridh2018precision} evaluated the performance of a handheld capacitance moisture meter and observed that the accuracy decreased for wood chips with MC$>$50\%, which could be a limitation of capacitance-based methods. Capacitance-based methods, although capable of detecting moisture changes in chip piles, suffer from errors due to their assumption of considering the wood chips as uniform material \citep{pan2016predicting}}. 

{\color{myblue}\citet{d2010simple} introduced a wood-chip humidity measurement technique utilizing time-domain-reflectometry (TDR). Their approach employs wire probe pulse signals to determine Round Trip Time (RTT), correlating with wood-chip humidity levels. Experimental and simulation results validate the method, demonstrating sensitivity to humidity changes and the potential for low-cost monitoring systems. However, microwave-based methods face challenges due to air gaps between chips \citep{d2010simple}; X-ray and NMR techniques are expensive \citep{barale2002use} and applicable only to small sample quantities \citep{hultnas2012determination}}. The Wi-Fi-based method \citep{suthar2021multiclass} is relatively new and has not been explored extensively. Machine learning and statistical modeling, particularly partial least squares (PLS) regression, are commonly used in these indirect methods. However, capturing the complex relationships between heterogeneous wood chips and the MC calls for a more robust and state-of-the-art technique.

Recent advancements in machine vision techniques, coupled with the superior computational ability of high-end computers and cutting-edge camera sensors, offer promising opportunities for employing machine vision-based methods in determining the MC of wood chips \citep{rahman2024comprehensive}. In this study, our objective is to investigate the applicability of an image-based method by leveraging deep learning and machine vision techniques. The study conducted by \citet{plankenbuhler2020image} focused on assessing wood chip quality attributes and mixture ratios using monochrome images and regression models. Their regression model employed hand-crafted features such as the brightness and textures of wood chips. 

In a rudimentary study, \citet{rahmaninterpretable} explored the use of a deep learning model for predicting the MC from images. {\color{myblue}However, their study was limited by a small dataset comprising only 30 images.} Consequently, in our study, we have developed a large-scale image dataset consisting of 1,600 RGB (Red Green Blue) images of wood chips. This dataset includes images collected from two different sources, one of which contains multiple batches, ensuring a diverse and representative dataset for our analysis. There exist several other image-based methods with different focuses related to wood chips, such as sorting wood chip types \citep{grigorev2021improving}, sorting wastes from wood chips \citep{wooten2011discrimination,verheyen2016vision}, and determining the size distribution of wood chips \citep{febbi2013determining, febbi2015automated}. However, the complete potential of the deep learning framework has yet to be explored for MC determination. {\color{myblue}If developed properly, the image-based deep learning model will be able to predict moisture content instantaneously and non-destructively. Such a machine vision-based strategy will circumvent the necessity of any time-consuming lab tests and expensive equipment. As a result, the moisture content determination in the wood chip industry will be near real-time. In the pursuit of developing high-performing deep neural architectures, we leveraged the Neural Architecture Search (NAS) and hyperparameter optimization methods. The developed neural networks, namely \textbf{MoistNet}, instantaneously predict the MC of wood chips from RGB images with higher accuracy and faster inference speed.}

The contributions of this study are highlighted as follows:
\begin{enumerate}

\item A carefully collected and annotated wood chip image dataset from different batches and sources is presented, establishing a strong foundation for wood chip MC class prediction through machine vision. 

\item Two high-performing neural networks are developed using Neural Architecture Search (NAS) and hyperparameter optimization: (a)  \textbf{MoistNetLite}, a lightweight model with comparable performance, and (b) \textbf{MoistNetMax}, a heavy model with high competitive accuracy.

\item {\color{myblue}The proposed MoistNet models outperform sixteen state-of-the-art deep learning models for multi-class MC prediction, demonstrating the potential for applicability as an industrial-scale moisture content predictor.}

\item MoistNet models demonstrate great robustness in a comprehensive sensitivity analysis when confronted with changes in the dataset, cross-validation settings, and several hyperparameters
including batch size, learning rate, optimizer, and weight initialization.

\end{enumerate}

The remaining part of this article is organized as follows: Section 2 focuses on the materials and methods, detailing the data collection strategy, MoistNet model development, and evaluation metrics. In Section 3, the experiments and results are presented, covering the experimental setup, training pipeline, performance results, and sensitivity analysis. Finally, Section 4 concludes the paper by summarizing the key findings and insights.

\section{Materials and Method}
In this section, we present the data acquisition process, the development of MoistNet models, a comprehensive list of baseline models, and the evaluation metrics used for performance assessment.
\begin{figure}[htbp]
  \centering
  \includegraphics[width=\textwidth]{ 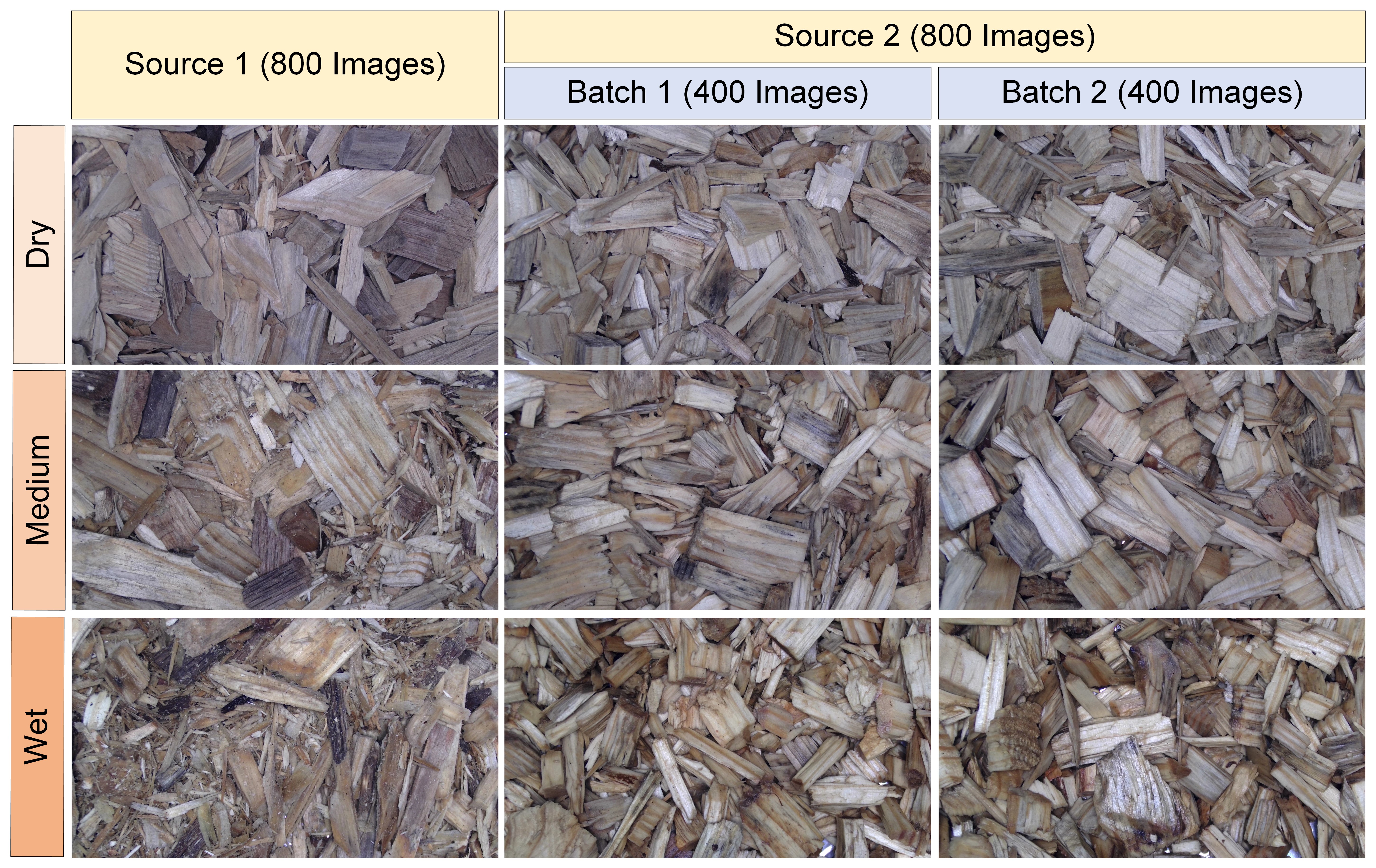}
  \caption{(Best viewed in color) Samples from the wood chip image dataset. Dry, medium, and wet represent three moisture classes where \textit{dry:} $\leq$15\%, \textit{medium:} 16-35\%, and \textit{wet:} $\geq$36\% moisture contents.}
  \label{fig:dataset_sample}
\end{figure}

\label{sec:methodology}
\subsection{Wood Chip Dataset Acquisition}
\subsubsection{Chip Sourcing}
\begin{figure}[htbp]
  \centering
  \includegraphics[width=0.75\textwidth]{ 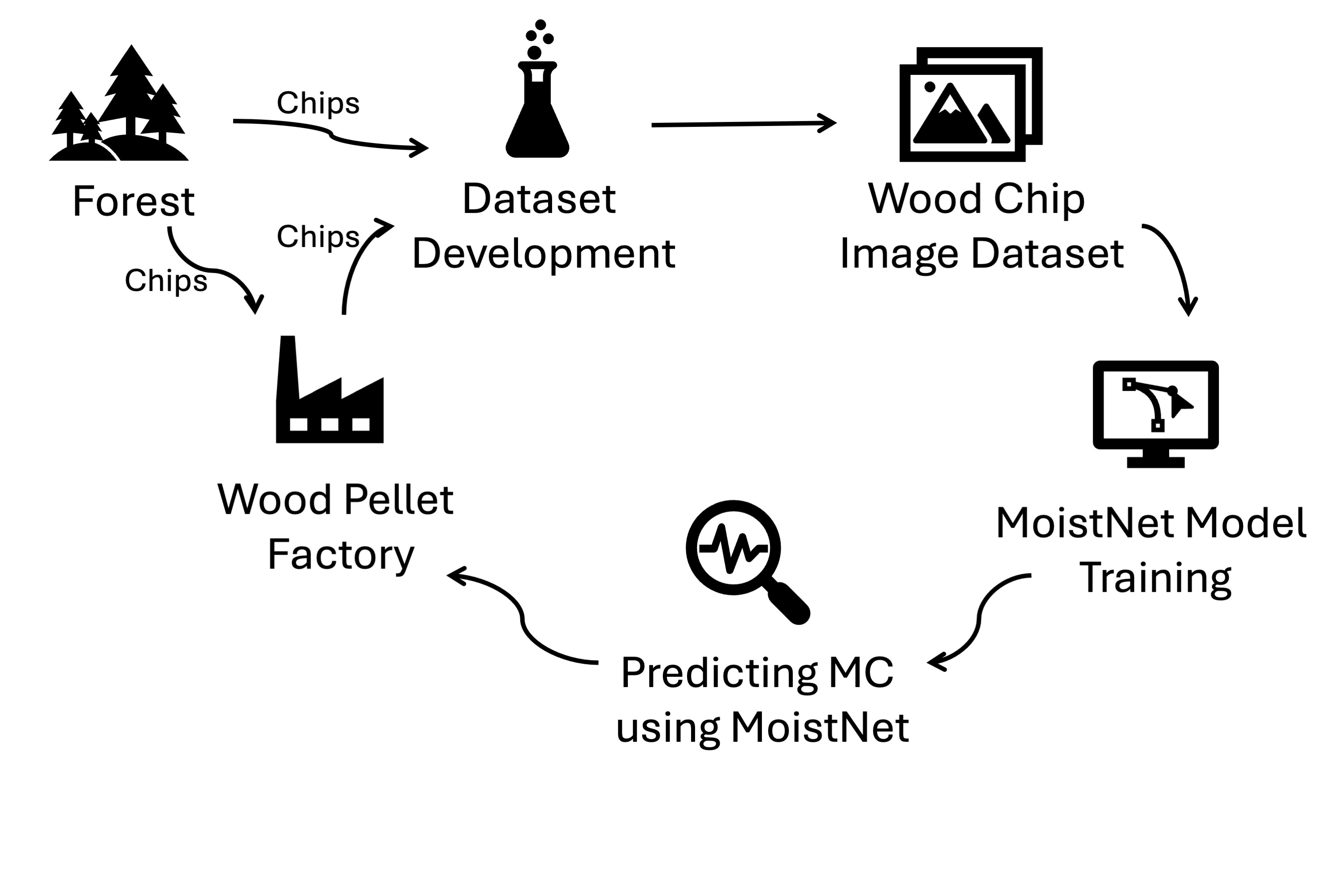}
  \caption{{\color{myblue}Overview of the proposed image-based wood chip moisture content class prediction method.}}
  \label{fig:process}
\end{figure}

Wood chip MC class prediction is an inherently challenging task due to the heterogeneity of the chips. Such heterogeneity could come from different perspectives, including plant type, chipping method, source of the chips, chip size, and so on. In this study, we collected wood chips from two different sources, as shown in Figure \ref{fig:dataset_sample}. While both sources were bio-fuel processing plants, the wood chips obtained from each source exhibited significant differences. Specifically, the chips from source 1 were gathered from the forest environment and referred to as \textit{inwood chips}. On the other hand, the chips from source 2 were collected from end-cuts of kiln-dried lumber from a lumber mill and referred to as \textit{lumber chips}. For source 2, we collected two distinct batches of lumber chips. It is worth noting that wood chips are typically stored in large piles \citep{kuptz2020fuel}. In our study, we collected chips from a sizable wood chip pile, but the collection points were approximately 50 meters apart. These distinct collections were considered batch 1 and batch 2, respectively. {\color{myblue}Figure \ref{fig:process} illustrates the overall process flow of the proposed image-based moisture content class prediction method.}

\subsubsection{Wood Chip Sample Preparation}
Raw wood chips exhibit a broad range of MC, which can vary depending on the location within the chip pile. In the center of the pile, the MC can significantly decrease, reaching as low as 25\% \citep{iwan2017influence}. Conversely, in the upper and outer parts, it can increase to 65-70\% compared to the initial MC \citep{iwan2017influence}. Since Deep Learning (DL) techniques require a diverse dataset encompassing various MC ranges, we artificially adjusted the MC of the chips first by completely drying them and then adding a certain amount of water. 

To ensure that the raw wood chips were completely dry, we utilized an oven dryer set at 105°C for 24 hours to remove the existing MC. Subsequently, a rotating mortar mixer was employed to effectively mix water with the chips. The amount of water to be added was calculated using Equation (\ref{eqn:water}). 
\begin{equation}
\label{eqn:water}
    m_{water} = m_{wet} \times mc\:\%
\end{equation} 
which was derived from $mc\% = 1 - m_{dry}/m_{wet}$, where $m_{water}=m_{wet}-m_{dry}$ represents the amount of water to be added, $m_{wet}$ and $m_{dry}$ indicate the wet weight and dry weight of the chips, respectively, and $mc\%$ corresponds to the intended MC. 
\begin{figure}[htbp]
  \centering
  \includegraphics[width=0.4\textwidth]{ 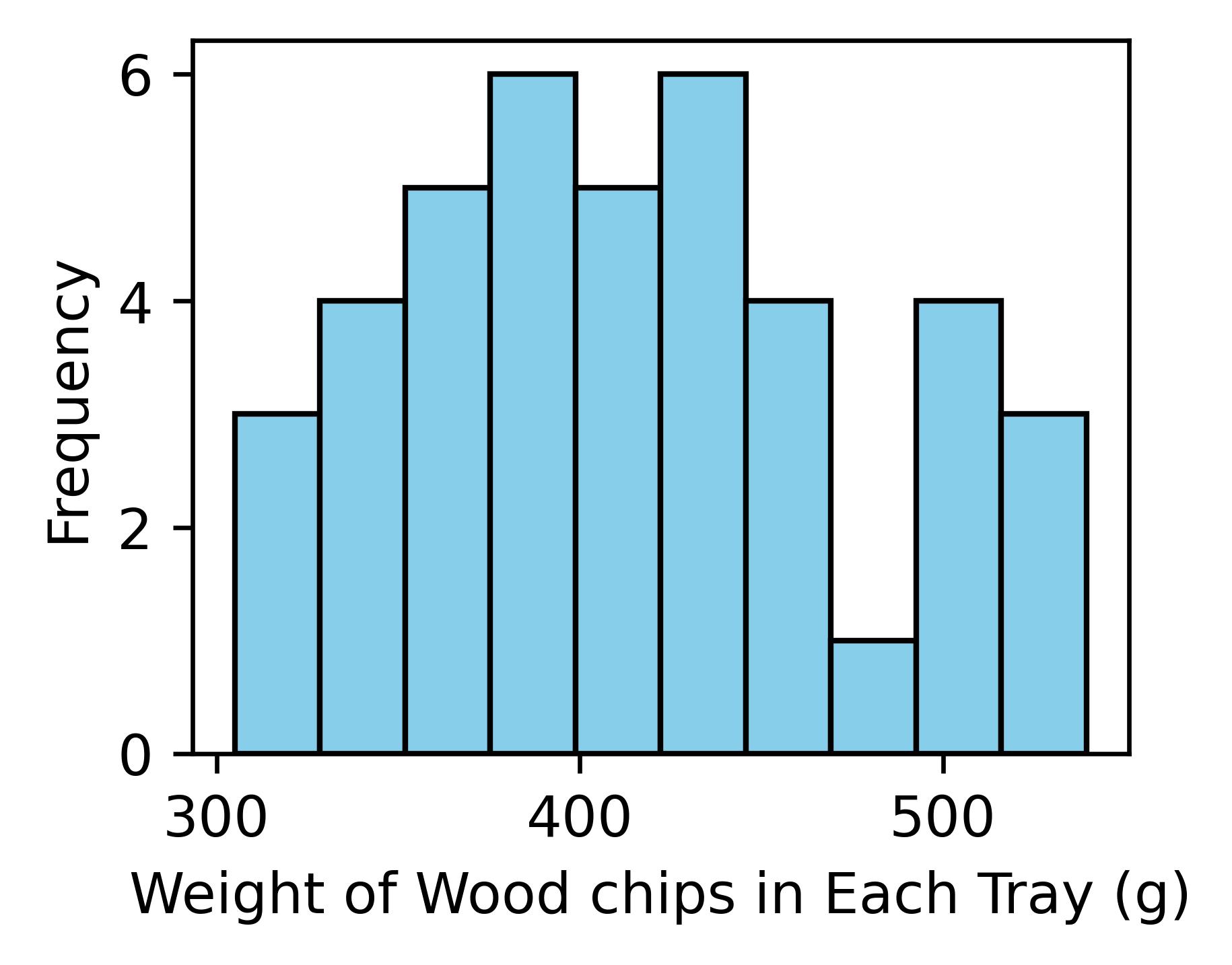}
  \caption{Distribution of weights of wood chips taken in each tray during image acquisition.}
  \label{fig:weight}
\end{figure}

Water was added to the dry chips using a high-volume, low-pressure paint sprayer, facilitating the atomization of the water to ensure uniform dispersion within the rotating mortar mixer. It is worth noting that during the mixing process, a certain amount of water content would evaporate, potentially affecting the final MC of the chips. To compensate for this evaporation, an additional 10-12\% of the calculated amount of water was added to the chips. This extra water accounted for the moisture loss during mixing. Once the mixing process was completed, the rotating mortar mixer was left running for an additional 15 minutes to ensure proper blending of the wood chips. This meticulous procedure guaranteed a consistent and well-mixed MC in the prepared chips.

\begin{figure}[htbp]
  \centering
  \includegraphics[width=\textwidth]{ 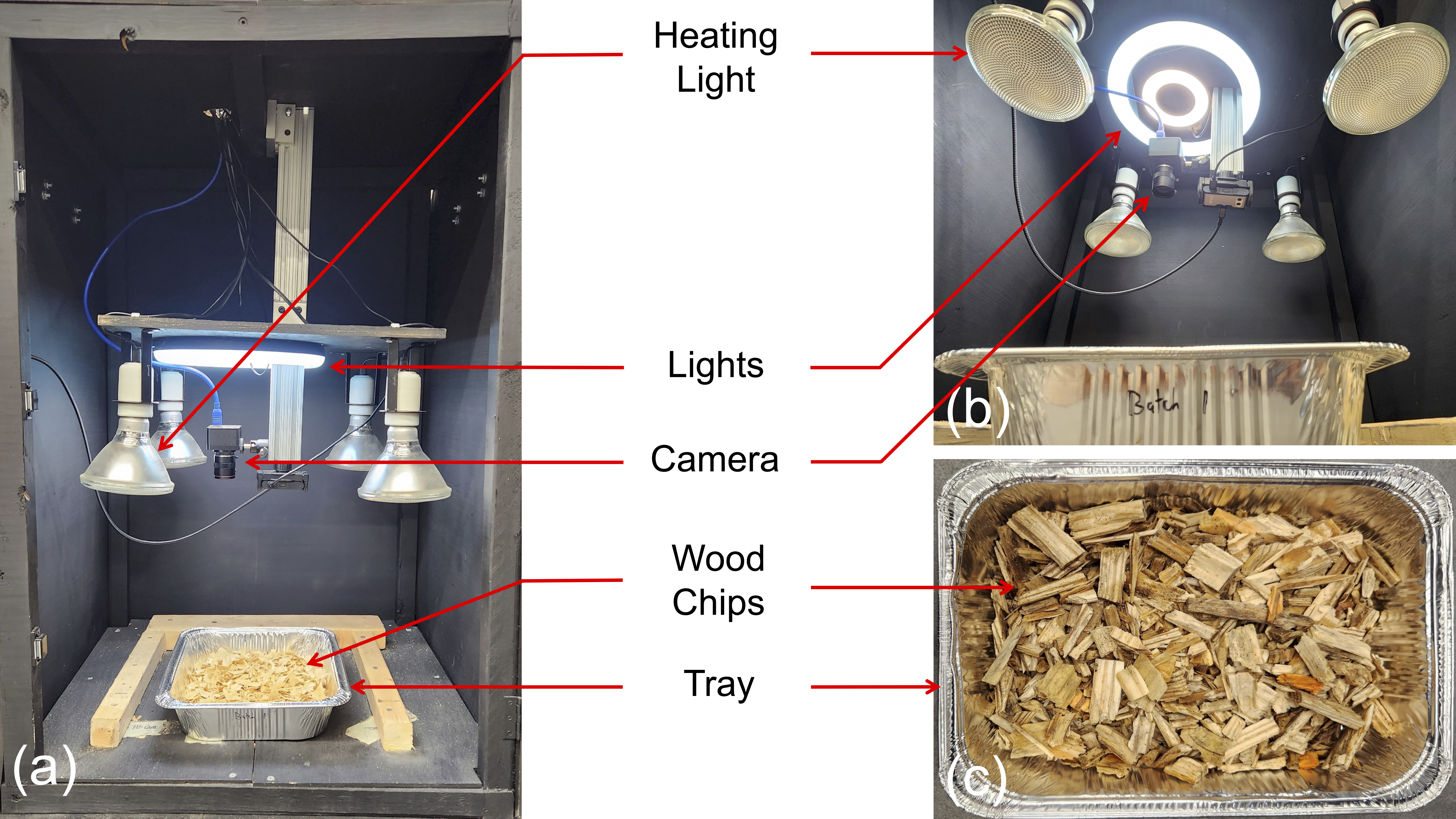}
  \caption{(Best viewed in color) (a) Data collection setup, (b) Uniform white lights, (c) Wood chips in the tray to capture images. Heating lights were not used in this experiment.}
  \label{fig:station}
\end{figure}

\subsubsection{Sample Image Acquisition}
 The wood chips were placed in foil trays with dimensions of 29.2 cm $\times$ 22.9 cm $\times$ 6.4 cm. Each tray contained wood chips weighing between 300-550 grams. Figure \ref{fig:weight} demonstrates the distribution of weights of wood chips taken in each tray for capturing RGB images. We have carefully designed a data collection station that consists of a closed box (with a door) as shown in Figure \ref{fig:station}(a). Two ring-shaped white lights were installed at the top of the box as illustrated in Figure \ref{fig:station}(b). We used such a closed box to ensure the same lighting condition for each sample. \citet{plankenbuhler2020image} used green lights for wood chip quality assessment. However, lighting color didn't affect their regression model because the evaluation was carried out on grey-scale images. An industrial camera (Hotpet 8MP USB Industrial Camera with Sony IMX179 Sensor) was positioned at the top of the box to capture images of the wood chips, as depicted in Figure \ref{fig:station}(a). The camera was set at a distance of 14 cm from the bottom surface of the box. Since the camera sensor did not possess auto-focusing capabilities, we manually adjusted the settings to ensure proper focus and clear image capture. Figure \ref{fig:station}(c) provides a visual representation of the wood chip samples as seen through the camera lens. 

During the image-capturing process, we ensured that the door of the box remained closed. We adopted a method of capturing 20 images from each tray of wood chips, shuffling the chips after each capture. This approach allowed us to capture the moisture features of the chips that were not initially on the surface. This is an important consideration since previous image-based methods \citep{plankenbuhler2020image} and near-infrared (NIR) spectra-based methods \citep{nascimbem2013determination, liang2019determination, amaral2020estimation, toscano2022performance} are limited in their ability to capture only the surface MC of the wood chips. By shuffling the chips, we were able to capture a more comprehensive representation of the moisture distribution throughout the samples. Figure \ref{fig:dataset_sample} illustrates the sample wood chip images captured in our data collection station.

\begin{figure}[tbp]
  \centering
  \includegraphics[width=0.8\textwidth]{ 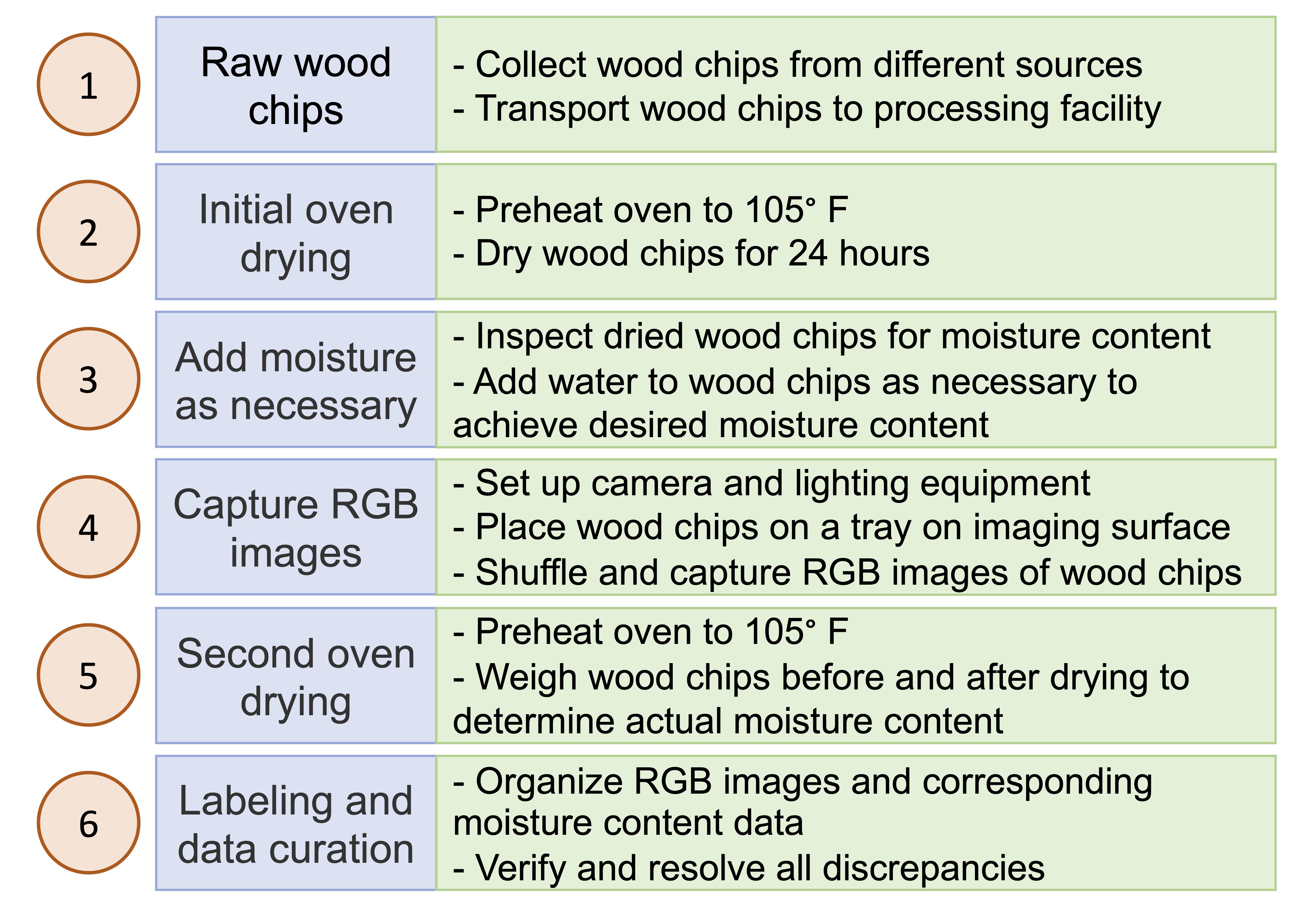}
  \caption{Wood chip image data collection process.}
  \label{fig:data_collection}
\end{figure}

\subsubsection{Data Labelling}
For any machine learning-based method of wood chip MC class prediction, data labeling is the most time-consuming step. The dependence on the conventional oven drying process to create ground truth labels is one of the major bottlenecks. In this study, we also had to rely on the oven-drying process to get the actual MC values. We measured the weight of the wood chip samples along with the tray when the image collection was complete. Then the samples were placed in an oven dryer pre-heated to 105°C for 24 hours. Finally, we measured the weight again after completion of the drying operation and calculated the MC of the samples using the Equation (\ref{eqn:mc}).
\begin{equation}
    \label{eqn:mc}
    mc \: (\%) = \frac{(m_{wet}-m_t)-(m_{dry}-m_t)}{(m_{wet}-m_t)} \times 100
\end{equation}
where, $m_{wet}$, $m_{dry}$, and $m_t$ refer to the weight of wood chips before drying, the same after drying, and the weight of the tray, respectively. We planned to prepare at least 10 different moisture levels each 5\% apart starting from 5\%. However, due to the heterogeneity of the chips and instrumental errors, we were able to generate moisture levels of 2\%, 10\%, 15\%, 20\%, 25\%, 26\%, 33\%, 39\%, 41\%, and 50\%. {\color{myblue}In this work, based on the discussion with the domain experts and requirements of the industry partners, we have created three classes from these moisture ranges such as \textit{dry:} $\leq$15\%, \textit{medium:} 16-35\%, and \textit{wet:} $\geq$36\%. However, we have also conducted experiments considering 5 classes of MC (see Section \ref{sec:experiment}). Figure \ref{fig:data_collection} summarizes all the steps of the wood chip data collection process adopted in this study. }

\subsection{Development of MoistNet}
\begin{figure}[htbp]
  \centering
  \includegraphics[width=0.8\textwidth]{ 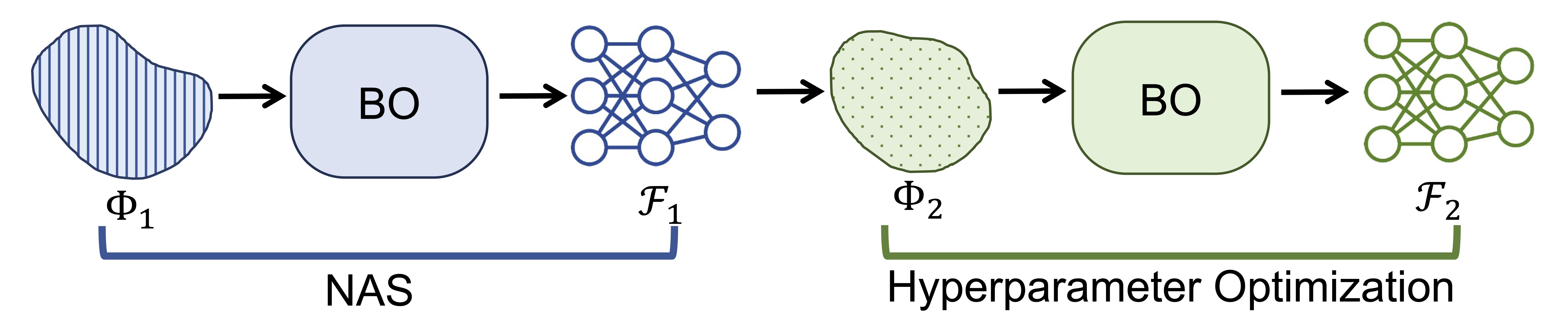}
  \caption{MoistNet development pipeline. In the first stage, NAS is employed to search for a better intermediate architecture $\mathcal{F}_1$ from the search space $\Phi _1$. In the second stage, hyperparameters such as learning rate and optimizers are optimized to achieve the best model architecture $\mathcal{F}_2$ from the refined search space $\Phi _2$. Both of these stages use the Bayesian Optimization (BO) algorithm as the search method.}
  \label{fig:nas}
\end{figure}

In this work, we employed Neural Architecture Search (NAS) and hyperparameter optimization using two search spaces to automate the development of the MoistNet model. NAS is the systematic process of automating the architecture engineering of deep learning models \citep{elsken2019neural}. NASNet \citep{zoph2018learning} is one such architecture that is optimized on the CIFAR10 \citep{krizhevsky2009learning} dataset and then applied to the ImageNet \citep{deng2009imagenet} dataset to achieve state-of-the-art results. Figure \ref{fig:nas} demonstrates the pipeline for MoistNet model development. 

A well-defined search space is vital for both NAS and hyperparameter optimization \citep{zoph2018learning,liu2017hierarchical}. We defined two search spaces namely $\Phi _1$ and $\Phi _2$ for NAS and hyperparameter optimization, respectively (see \ref{sec:appendix_a} and \ref{sec:appendix_b}). $\Phi _1$ was initialized with the normalization layer, augmentation layer, and feature extraction block type. On top of these, we added the classification layer that involved the choice of spatial reduction type and several dropout options. The spatial reduction types included different transformations of the features including flattening, taking the global average, or global maximum. The search space also included the choice of the optimizer, and learning rate. Moreover, the augmentation layer had several options: translation, horizontal flip, vertical flip, rotation, zoom, and contrast. The most important choice in this search space is the choice of feature extraction block. The options for this block include a vanilla architecture with several convolutional, max-pooling, and dropout layers; several ResNet architectures \citep{he2016deep,he2016identity}; an Xception architecture \citep{chollet2017xception}; and several EfficientNet architectures \citep{tan2019efficientnet}. While the vanilla architectures result in a model with a very low number of model parameters, other architectures yield models having a higher number of parameters. We denote the neural architecture generated in this stage as $\mathcal{F}_1$ having the architecture configuration $C^*$.

The NAS is a powerful tool to find a better model for any particular dataset. However, the search space $\Phi _1$ has some limitations. For example, the options for optimizers and the learning rate are not comprehensive. There exist several other advanced optimizers such as \verb|RMSProp|, \verb|Adagrad|, \verb|Adadelta|, \verb|Nadam|, etc. Moreover, the list of learning rates only included several discrete values leading to a search space that might miss the optimal learning rate for the respective model. A fine-grained search space of the learning rate could potentially lead to better optimization of the learning rate. Therefore, we conducted the second level of hyperparameter optimization using the Bayesian optimization method on a refined search space $\Phi _2$ including the optimizers, learning rate, and dropout rate (see \ref{sec:appendix_b} for the search space $\Phi _2$). Therefore, the MoistNet model development process involved two stages. In the first stage, we explore the search space $\Phi _1$ to discover an initial optimal neural architecture $\mathcal{F}_1$. Once this architecture is obtained, we move on to the second stage, where we conduct further hyperparameter optimization using the search space $\Phi _2$. This two-stage approach would lead to the development of a more refined and better-optimized model $\mathcal{F}_2$ having hyperparameter configuration $H^*$, which we refer to as MoistNet.

The overall goal of NAS and hyperparameter optimization through the Bayesian Optimization (BO) method is to find the optimal architecture configuration $C^*$ and hyperparameter configuration $H^*$ that maximizes the objective functions: $f_1$ and  $f_2$.
Therefore, we formulate the bi-level optimization problem as:\\
Step-1: NAS\\
\begin{equation}
\label{eqn:obj1}
    \arg\max_{C_{lik_i}} f_1(C_{lik_i}, X, y)
\end{equation}
subject to\\
\begin{equation}
\label{eqn:cons11}
    \sum_{i=1}^{n}\sum_{k_i=1}^{G_i}C_{lik_i} = 1 , \quad \forall \quad l
\end{equation}
\begin{equation}
\label{eqn:cons12}
    C_{lik_i} \in \{0,1\}, \quad \forall \quad l, i, k_i 
\end{equation}
\begin{equation}
\label{eqn:cons13}
    k_i \in N, \quad \forall \quad i 
\end{equation}
where $C_{lik_i}$ denotes the $i$-th operation with $k_i$-th setting is selected at $l$-th layer if  $C_{lik_i}=1$, otherwise $C_{lik_i}=0$. $G_i$ is the total number of setting options for the $i$-th operation. For instance, \verb|translation_factor| is an operation in the \verb|augment| layer with two setting choices of 0.0 and 0.1.\\
Step-2: Hyperparameter Optimization\\
\begin{equation}
\label{eqn:obj2}
    \arg\max_{H_{ik}\in \Phi_2} f_2(H_{ik}, C^*, X, y)
\end{equation}
where $H_{ik}$ is the decision variable that represents the $i$-th operation with $k_i$-th setting is selected if  $H_{ik}=1$, otherwise $H_{ik}=0$. In this study, test accuracy has been adopted as the objective function $f_1$ and $f_2$ to guide BO in Equation (\ref{eqn:obj1}) and (\ref{eqn:obj2}). Equation (\ref{eqn:cons11}),(\ref{eqn:cons12}), (\ref{eqn:cons13}) represents the search space constraints. 

\subsection{Baseline Models}
\label{baselines}
We evaluated the performance of our proposed MoistNet models by comparing them with sixteen state-of-the-art deep image classification models. The baseline models were grouped into different categories based on their architectural characteristics:

\begin{enumerate}

\item ResNet \citep{he2016deep,he2016identity}: ResNet models are known for their deep architecture and residual connections. They include ResNet50, ResNet50V2, ResNet101, ResNet101V2, ResNet152, and ResNet152V2, which offer varying depths for image classification tasks.

\item Inception \citep{szegedy2016rethinking, szegedy2017inception}: Inception models, including InceptionV3 and InceptionResNetV2, employ inception modules that enable efficient multi-level feature extraction by combining filters of different sizes.

\item MobileNet \citep{howard2017mobilenets}: MobileNet is a lightweight deep-learning model designed specifically for mobile and embedded devices. It achieves a good balance between accuracy and computational efficiency.

\item DenseNet \citep{huang2017densely}: DenseNet models, including DenseNet121, DenseNet169, and DenseNet201, employ densely connected convolutional layers, enabling feature reuse and strong gradient flow throughout the network.

 \item Xception \citep{chollet2017xception}: Xception is an extension of the Inception architecture that replaces the traditional inception modules with depthwise separable convolutions, enabling more efficient and effective feature extraction.

\item EfficientNet \citep{tan2019efficientnet}: EfficientNet models, such as EfficientNetB0, EfficientNetB1, and EfficientNetB2, leverage compound scaling to achieve state-of-the-art performance by balancing model depth, width, and resolution for efficient and effective feature representation.

\end{enumerate}

\subsection{Evaluation Metrics}
We employed a set of evaluation metrics to evaluate the performance of the baselines and MoistNet, focusing on classification performance and computational efficiency. Accuracy, precision, recall, and F1-score have been used to measure classification performance. These metrics were calculated using Equations (\ref{eqn:acc}) - (\ref{eqn:f1}), where TP, TN, FP, and FN represent true positive, true negative, false positive, and false negative, respectively. 
\begin{equation}
\label{eqn:acc}
    Accuracy = \frac{TP + TN}{TP + TN + FP +FN}
\end{equation}
\begin{equation}
\label{eqn:pre}
    Precision = \frac{TP}{TP + FP}
\end{equation}
\begin{equation}
\label{eqn:rec}
    Recall = \frac{TP}{TP + FN}
\end{equation}
\begin{equation}
\label{eqn:f1}
    F1\: score = \frac{2\times Precision \times Recall}{Precision + Recall}
\end{equation}

On the contrary, the computational efficiency of the model has been evaluated comprehensively based on several factors, including training time, inference time, and number of parameters in the model. For training time evaluation, we calculated the average time taken by each model to complete the training process using 4-fold cross-validation. Notably, we implemented an early stopping criterion to halt the training process when no improvement in validation accuracy was observed for 20 consecutive epochs. Inference time was determined by averaging the time required to make predictions on the test images. Lastly, the number of parameters in the model served as an indicator of its complexity and memory requirements.

\section{Experiments and Results}
\label{sec:experiment}
In this section, we illustrated the experimental setup for NAS, hyperparameter optimization, the training pipeline for the MoistNet model, and the baseline models. We discussed the results from different viewpoints and performed an extensive sensitivity analysis to gain valuable insights into the MoistNet models. {\color{myblue}Finally, we discuss the broader environmental, economic, or logistical implications of the proposed deep learning-based MC class prediction method for industry applications.} 
\subsection{NAS and Hyperparameter Optimization}
\begin{figure}[tbp]
  \centering
  \begin{subfigure}[b]{.3\textwidth}
    \centering
    \includegraphics[height=0.56\textwidth, width =\textwidth]{ 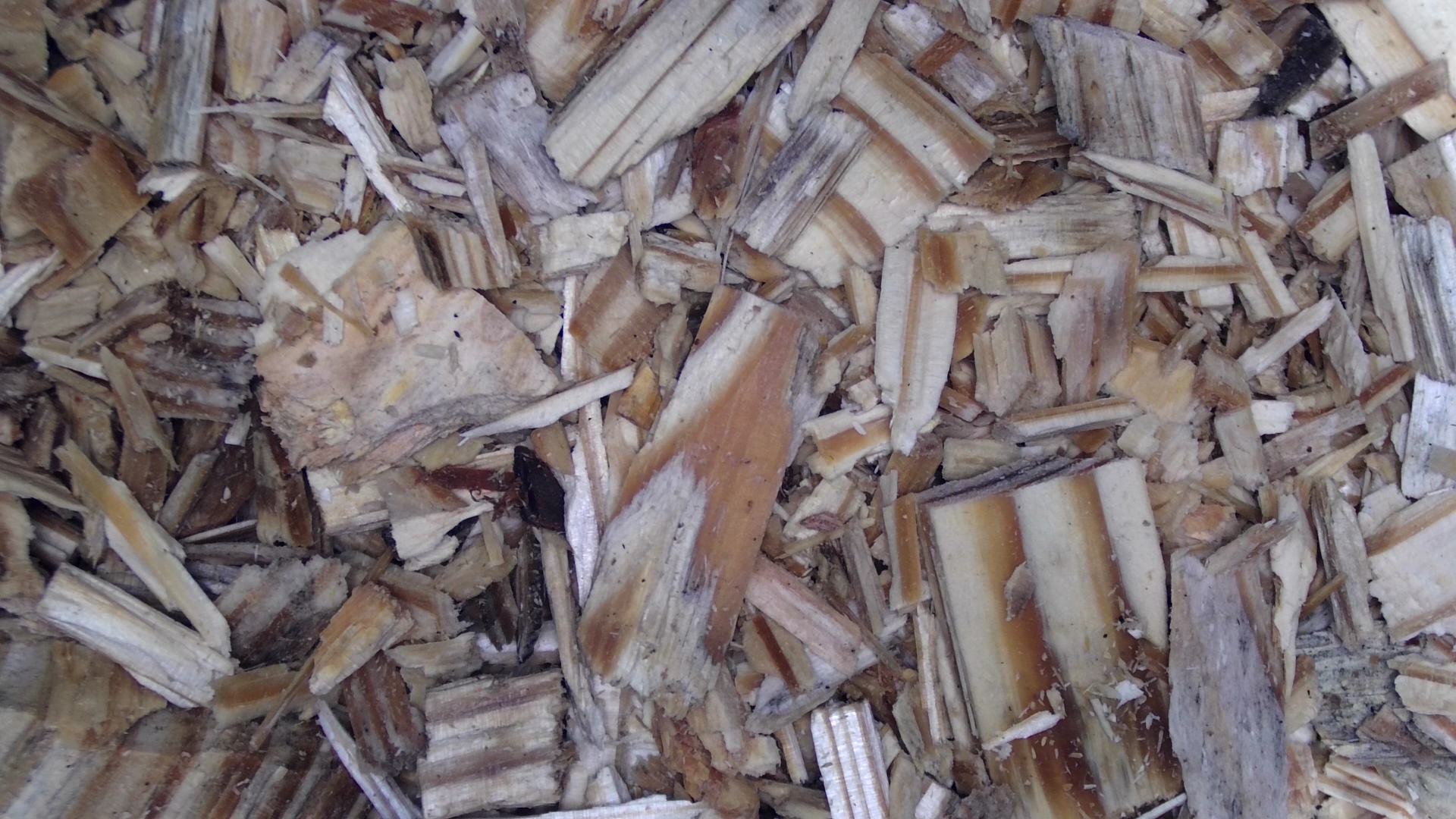}
    \caption{Raw image sample ($1920\times1080$ pixels)}
  \end{subfigure}
  \begin{subfigure}[b]{.5\textwidth}
    \centering
    \includegraphics[height=0.35\textwidth, width =\textwidth]{ 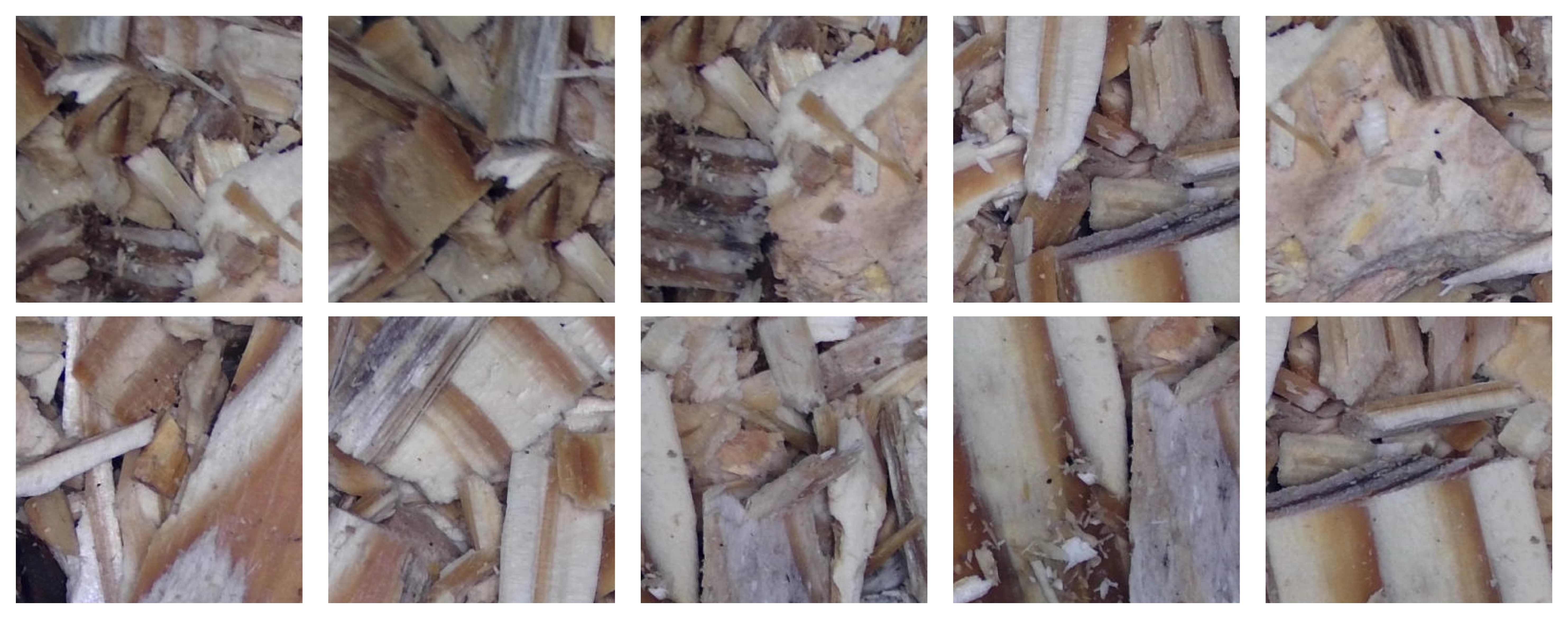}
    \caption{Generated random patch images ($300\times300$ pixels each)}
  \end{subfigure}
  \caption{(Best viewed in color) Random patch generation from a large image of wood chips to enhance the number of training samples. Ten patch images were generated from each large image.}
  \label{fig:patch_sample}
\end{figure}

We collected a total of 1600 images of the wood chip samples, including two sources, with each source contributing 800 images (see Figure \ref{fig:dataset_sample}). Additionally, within source 2, there were two batches, with each batch containing 400 images. To accommodate the data requirements of deep learning models, we adopted a random patch-generation strategy. This involved generating 10 small random patches, each measuring $300\times300$ pixels, from the original raw image, which had dimensions of $1920\times1080$ pixels. Figure \ref{fig:patch_sample} showcases an example of a raw image and the resulting random patches generated. Through this patch generation approach, we effectively increased the number of images by a factor of 10. {\color{myblue}Patch images were further resized to $224\times224$ pixels before using them in the training process as input to the models.}

\begin{figure}[htbp]
  \centering
  \includegraphics[width=\textwidth]{ 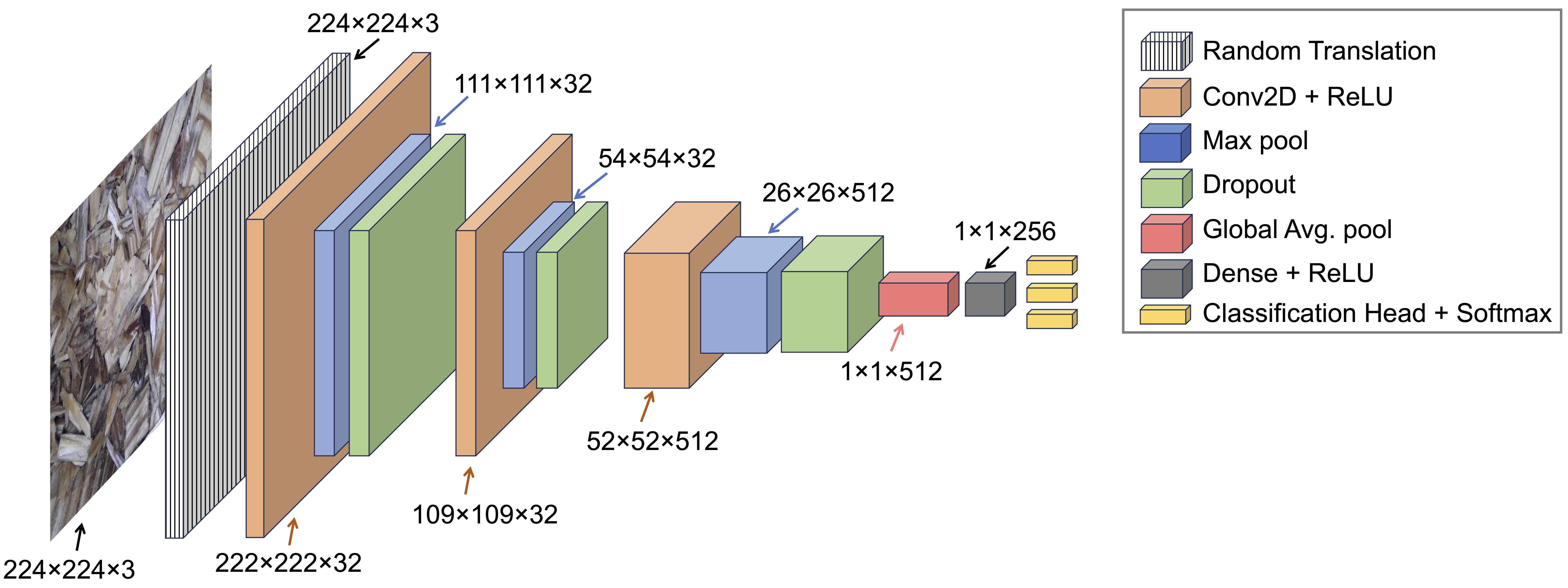}
  \caption{(Best viewed in color) MoistNetLite Architecture}
  \label{fig:moistnetlite}
\end{figure}

To develop the MoistNet architecture, we employed the \verb|AutoKeras|\footnote{ \url{https://autokeras.com/}} module to perform NAS as illustrated in Equations (\ref{eqn:obj1})-(\ref{eqn:cons13}), and then we utilized \verb|skopt|\footnote{ \url{https://scikit-optimize.github.io/stable/index.html}} to apply Bayesian optimization for hyperparameter optimization as illustrated in Equation (\ref{eqn:obj2}). {\color{myblue}We opted for a data splitting strategy of 0.75:0.25 ratio for training and testing, respectively, both during the NAS and hyperparameter optimization stages. Although adopting k-fold cross-validation would ensure better generalization of the developed model, we had to use specific splits (0.75:0.25) to reduce the run time of the optimization process. It is worth mentioning that certain models during the NAS failed to achieve satisfactory prediction accuracy. To avoid training these models further, we implemented an early stopping callback, which halted the training process for models that failed to converge toward better results within the initial 3 epochs. We used 20\% of the training set as the separate validation set, which guided the early stopping callback \citep{xu2018splitting}. Each potential neural architecture was trained for 10 epochs and then evaluated for the classification performance on the validation set. This validation accuracy is then used in the Bayesian optimization algorithm to guide the NAS. We ran the NAS for 1,500 iterations to explore a diverse range of architectures, which yielded a better neural architecture $\mathcal{F}_1$ based on the search space $\Phi _1$.}

\begin{figure}[htbp]
  \centering
  \includegraphics[width=0.8\textwidth]{ 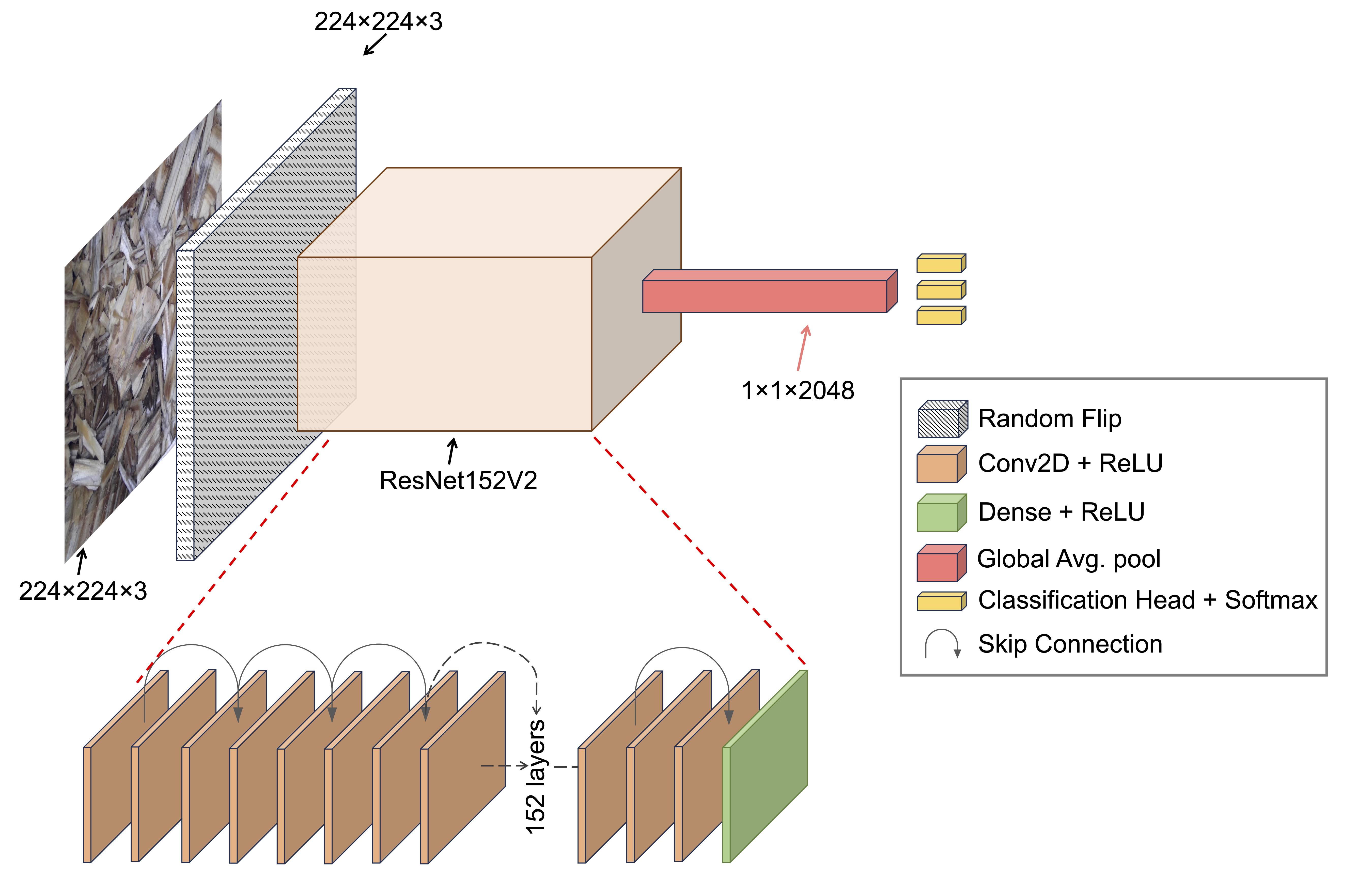}
  \caption{(Best viewed in color) MoistNetMax Architecture}
  \label{fig:moistnetmax}
\end{figure}

Next, we used this $\mathcal{F}_1$ architecture to fine-tune its hyperparameter with the search space $\Phi _1$. While the data split remained the same as NAS, in this case, we trained each architecture for 50 epochs and 500 iterations of the Bayesian optimization method. This increased number of training epochs allowed each architecture to explore the loss landscape extensively. However, the early stopping callback was used in this case but with 20 epochs as the patience parameter. This hyperparameter optimization step yielded the optimized architecture $\mathcal{F}_2$. 

\begin{figure}[htbp]
  \centering
  \includegraphics[width=\textwidth]{ 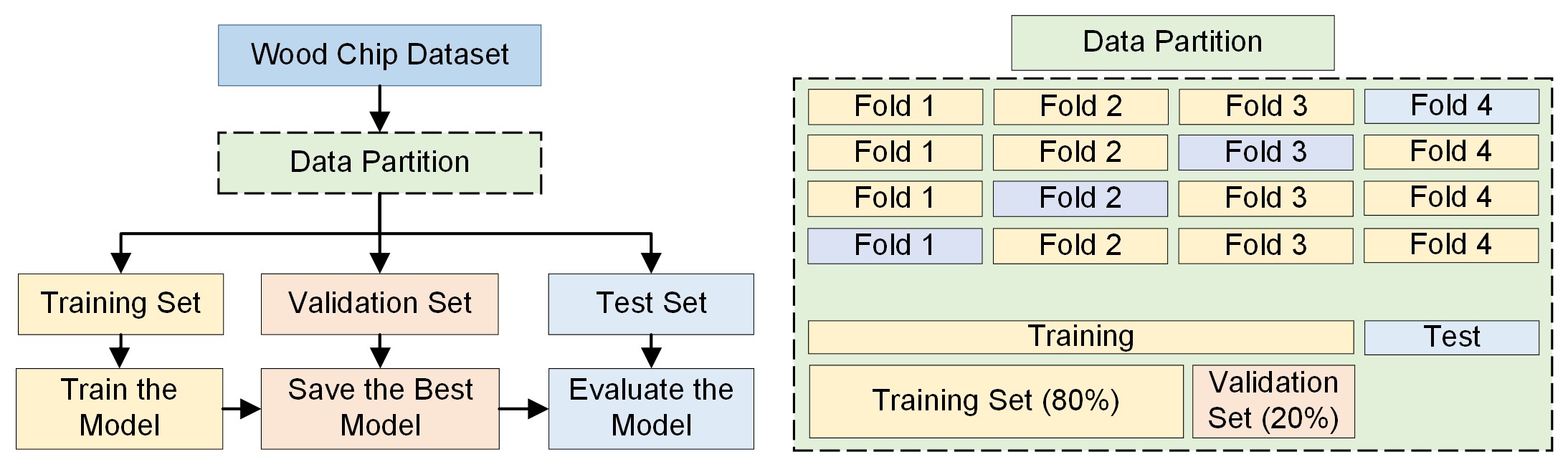}
  \caption{(Best viewed in color) Baseline model and MoistNet training pipeline.}
  \label{fig:training pipeline}
\end{figure}
\subsection{Performance of MoistNet and Baselines}
Employing the NAS and hyperparameter optimization, we introduced two neural networks: MoistNetLite and MoistNetMax. Figure \ref{fig:moistnetlite} and Figure \ref{fig:moistnetmax} illustrate the architectures of the MoistNetLite and MoistNetMax, respectively. {\color{myblue}MoistNetLite is a lightweight architecture comprised of an image translation layer, three consecutive layers of convolution, max pooling, and dropout, and a global average pooling layer followed by a dense layer. On the other hand, MoistNetMax is a deeper network with ResNet152V2 architecture as the backbone. In the MoistNetMax model, the input layer is followed by a random flip layer, the backbone (ResNet152V2), and then a global average pooling layer. Both MoistNetLite and MoistNetMax have a classification layer with a softmax activation function at the top.} {\color{myblue}MoistNetLite demonstrated comparable performance with minimal computational overhead (14 times less number of parameters and two times faster inference compared to the fastest baseline MobileNet). On the contrary, MoistNetMax exhibits exceptional precision at the expense of increased computational complexity. MoistNetMax achieved 9.6\% additional precision compared to the highest performing baseline ResNet152V2.}

We trained each baseline along with the MoistNet models for 200 epochs with an early stopping callback to halt the training if no improvement in validation accuracy was observed for 20 consecutive epochs. We employed a 4-fold cross-validation strategy to evaluate the performance. The reason behind the value of $K=4$ is explained later in Section \ref{sec:cross}. Following this strategy, we used three folds for training and the remaining one-fold for testing. The training samples were again divided into 0.8:0.2 proportions to get the training and validation sets. Figure \ref{fig:training pipeline} demonstrates the training pipeline and data partition strategy. During training, the model that performed the best on the validation set was saved and after the completion of the training, the saved model was used to evaluate the performance on the test set. {\color{myblue}We have used several data augmentation techniques to increase the robustness of the model, including random rotation, translation, zoom, and vertical and horizontal flip. }

\begin{sidewaystable}
\centering
\caption{Comparison of the performance of state-of-the-art deep learning models with MoistNet models in wood chip MC class prediction on the dataset from source 1. {\color{myblue}Each entry represents the mean and standard deviation from the 4-fold cross-validation strategy.} }
\label{tab:source1}
\resizebox{\textwidth}{!}{%
\begin{tabular}{@{}llllllllll@{}}
\hline
Model Backbone & Variants          & Loss        & Accuracy    & Precision   & Recall      & F1-score    & Training Time & {\color{myblue}\begin{tabular}[c]{@{}c@{}}Inference\\ Time (ms)\end{tabular}}  & {\color{myblue}\begin{tabular}[c]{@{}c@{}}\#Parameters\\ (million)\end{tabular}}  \\ \hline
Xception       &                   & 0.54 ± 0.18 & 0.77 ± 0.08 & 0.77 ± 0.08 & 0.76 ± 0.08 & 0.76 ± 0.08 & 44m 5s        & 2.90 ± 0.15    & 22.96        \\\hline
MobileNet      &                   & 0.42 ± 0.05 & 0.80 ± 0.04 & 0.81 ± 0.04 & 0.80 ± 0.04 & 0.80 ± 0.04 & 35m 39s       & 1.16 ± 0.03    & 4.28         \\\hline
\multirow{6}{*}{ResNet}       & ResNet50       & 0.37 ± 0.05 & 0.83 ± 0.03 & 0.83 ± 0.03 & 0.83 ± 0.03 & 0.83 ± 0.03 & 38m 18s & 2.93 ± 0.04 & 25.69 \\
               & ResNet50V2        & 0.40 ± 0.14 & 0.79 ± 0.10 & 0.79 ± 0.10 & 0.79 ± 0.11 & 0.79 ± 0.10 & 40m 28s       & 2.51 ± 0.04    & 25.67        \\
               & ResNet101         & 0.52 ± 0.15 & 0.74 ± 0.08 & 0.74 ± 0.08 & 0.74 ± 0.08 & 0.74 ± 0.08 & 43m 33s       & 5.04 ± 0.14    & 44.76        \\
               & ResNet101V2       & 0.41 ± 0.09 & 0.81 ± 0.04 & 0.81 ± 0.04 & 0.81 ± 0.04 & 0.81 ± 0.04 & 1h 11m 59s    & 4.45 ± 0.05    & 44.73        \\
               & ResNet152         & 0.39 ± 0.04 & 0.83 ± 0.02 & 0.83 ± 0.02 & 0.82 ± 0.02 & 0.83 ± 0.02 & 2h 3m 50s     & 7.45 ± 0.13    & 60.47        \\
               & ResNet152V2       & 0.36 ± 0.06 & 0.84 ± 0.04 & 0.84 ± 0.04 & 0.83 ± 0.04 & 0.83 ± 0.04 & 1h 58m 2s     & 6.74 ± 0.28    & 60.43        \\\hline
\multirow{2}{*}{Inception}    & InceptionV3    & 0.39 ± 0.07 & 0.83 ± 0.03 & 0.84 ± 0.03 & 0.83 ± 0.03 & 0.83 ± 0.03 & 40m 57s & 2.54 ± 0.07 & 23.9  \\
               & InceptionResNetV2 & 0.55 ± 0.11 & 0.77 ± 0.04 & 0.77 ± 0.04 & 0.75 ± 0.04 & 0.76 ± 0.04 & 1h 5m 24s     & 6.60 ± 0.28    & 55.91        \\\hline
\multirow{3}{*}{DenseNet}     & DenseNet121    & 0.63 ± 0.36 & 0.75 ± 0.11 & 0.75 ± 0.11 & 0.73 ± 0.12 & 0.74 ± 0.11 & 39m 1s  & 3.59 ± 0.17 & 8.09  \\
               & DenseNet169       & 0.60 ± 0.29 & 0.74 ± 0.12 & 0.75 ± 0.11 & 0.74 ± 0.13 & 0.74 ± 0.12 & 53m 59s       & 4.72 ± 0.06    & 14.35        \\
               & DenseNet201       & 0.49 ± 0.12 & 0.79 ± 0.04 & 0.80 ± 0.04 & 0.79 ± 0.04 & 0.79 ± 0.04 & 1h 6m 12s     & 6.09 ± 0.46    & 20.29        \\\hline
\multirow{3}{*}{EfficientNet} & EfficientNetB0 & 0.62 ± 0.19 & 0.69 ± 0.08 & 0.70 ± 0.08 & 0.68 ± 0.10 & 0.68 ± 0.09 & 58m 2s  & 2.55 ± 0.02 & 5.36  \\
               & EfficientNetB1    & 0.64 ± 0.15 & 0.65 ± 0.08 & 0.67 ± 0.08 & 0.58 ± 0.13 & 0.61 ± 0.11 & 48m 1s        & 3.48 ± 0.04    & 7.89         \\
               & EfficientNetB2    & 0.89 ± 0.47 & 0.58 ± 0.10 & 0.59 ± 0.11 & 0.45 ± 0.16 & 0.49 ± 0.14 & 39m 10s       & 3.63 ± 0.10    & 9.21         \\\hline
\multirow{2}{*}{MoistNet (Ours)}     & MoistNetLite   & 0.31 ± 0.11 & 0.87 ± 0.05 & 0.87 ± 0.05 & 0.87 ± 0.05 & 0.87 ± 0.05 & \textbf{6m 41s}  & \textbf{0.58 ± 0.01} & \textbf{0.29}  \\
               & MoistNetMax       & \textbf{0.24 ± 0.04} & \textbf{0.91 ± 0.02} & \textbf{0.91 ± 0.02} & \textbf{0.91 ± 0.02} & \textbf{0.91 ± 0.02} & 1h 15m 44s    & 6.56 ± 0.31    & 58.34        \\ \hline
\end{tabular}%
}
\end{sidewaystable}

\begin{figure}[htbp]
  \centering
  \begin{subfigure}[b]{0.45\textwidth}
    \centering
    \includegraphics[width=\textwidth]{ 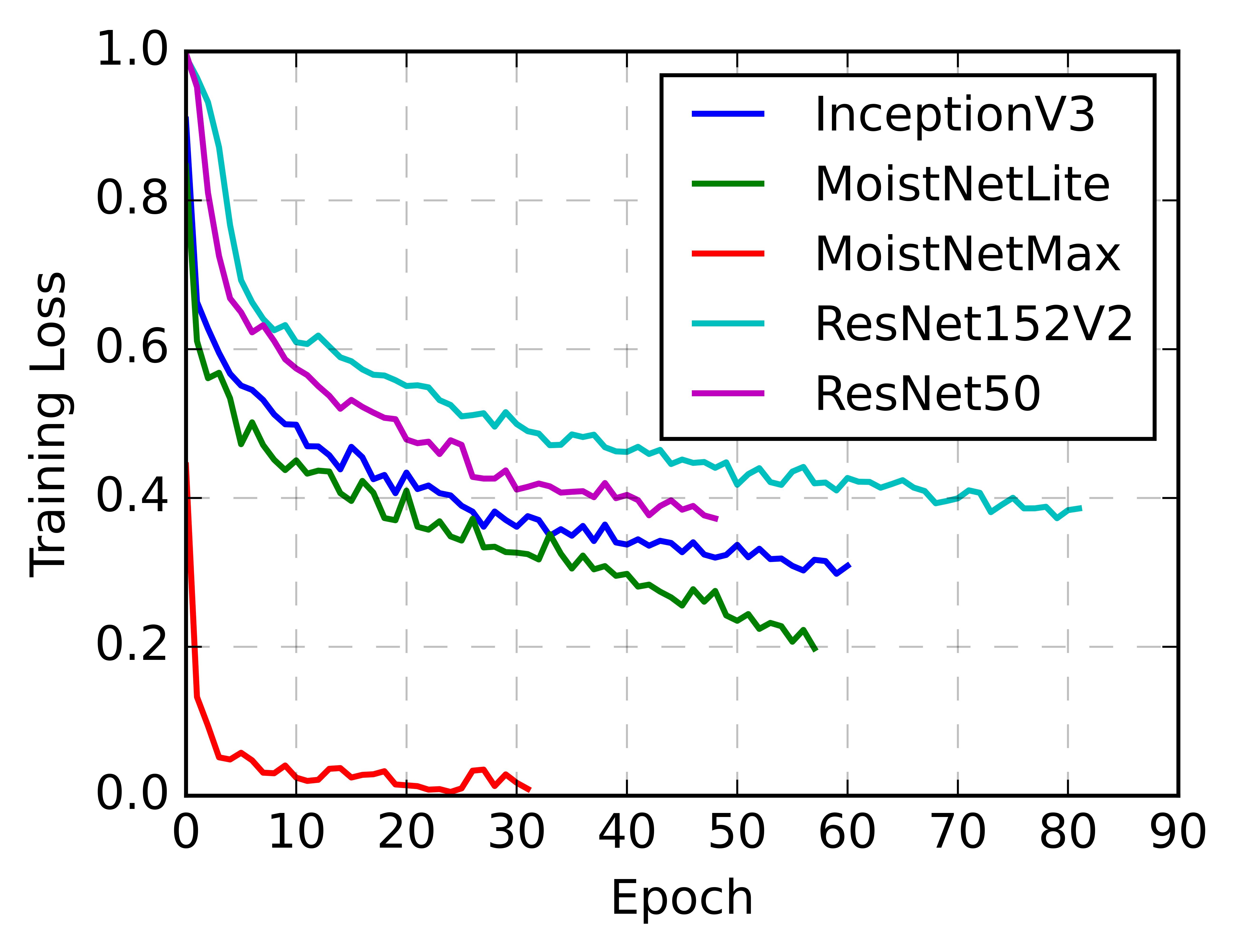}
    \caption{Training Loss}
  \end{subfigure}
  \begin{subfigure}[b]{0.45\textwidth}
    \centering
    \includegraphics[width=\textwidth]{ 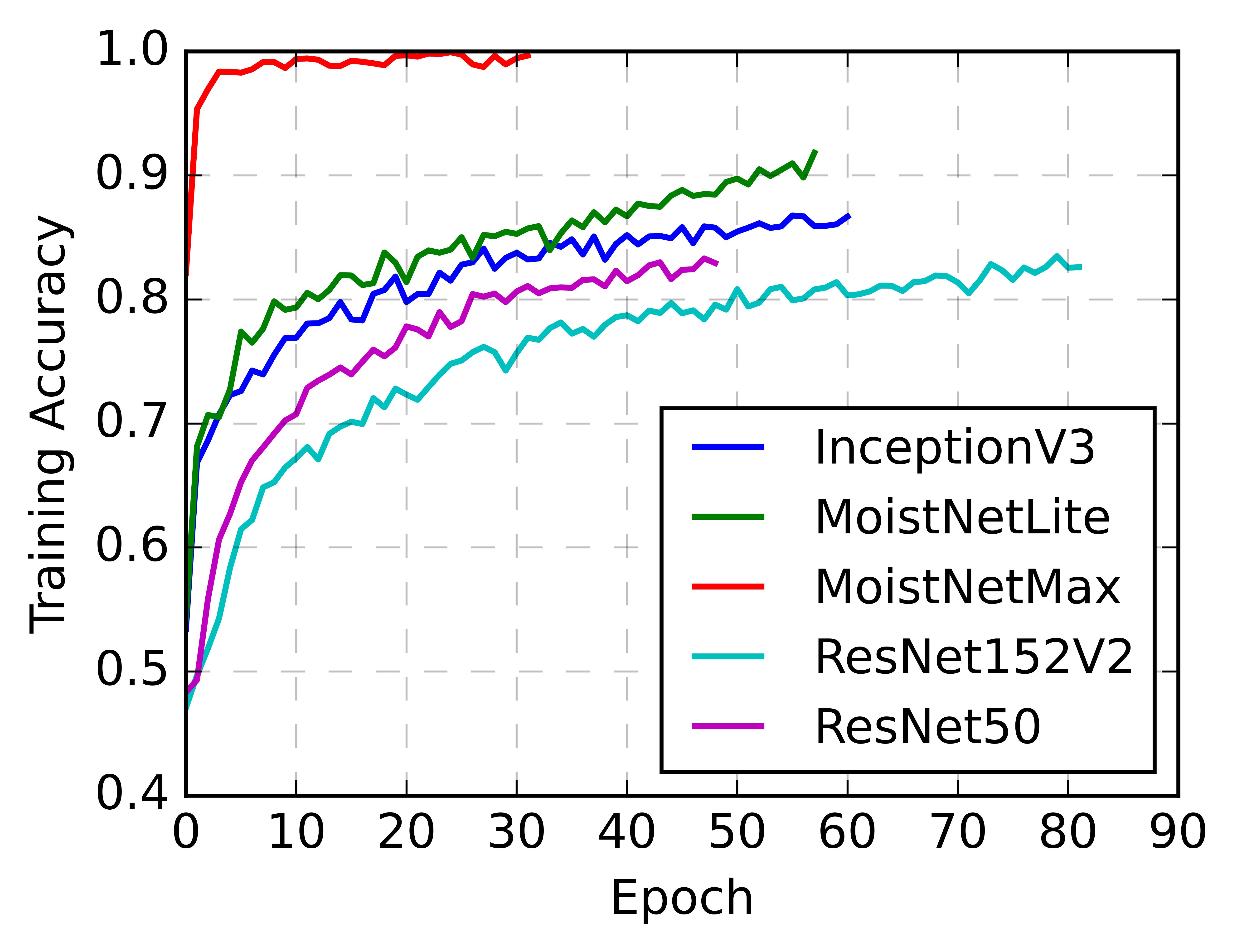}
    \caption{Training Accuracy}
  \end{subfigure}
  \caption{(Best viewed in color) Training curves for the top five models, including MoistNet models. Each model was trained with an early stopping criteria to stop training if no improvement in validation accuracy was observed for 20 consecutive epochs.}
  \label{fig:training_curves}
\end{figure} 
We used a batch size of 16, optimizer Stochastic Gradient Descent (SGD) \citep{bottou2010large} with a learning rate of 0.0001 and momentum of 0.9. All the models were trained from scratch by initiating the weight randomly. However, a random seed was selected to make the results reproducible. For the baseline models, we removed the top Imagenet classification head and added a global average pooling layer along with a fully connected layer and a classification layer for the wood chip MC classes. We recorded the training loss and training accuracy for all the models. Figure \ref{fig:training_curves} illustrates the training loss and training accuracy for the top five models, including MoistNet models. MoistNetLite appeared to converge faster than the three baseline models (ResNet50, ResNet152V2, and InceptionV3). On top of that, MoistNetMax exhibited exceptional performance in converging very fast. Table \ref{tab:source1} demonstrates the performance of all the baseline and MoistNet models on the dataset from source 1 with all the performance metrics. 

\begin{figure}[htbp]
  \centering
  \begin{subfigure}[b]{0.3\textwidth}
    \centering
    \includegraphics[width=\textwidth]{ 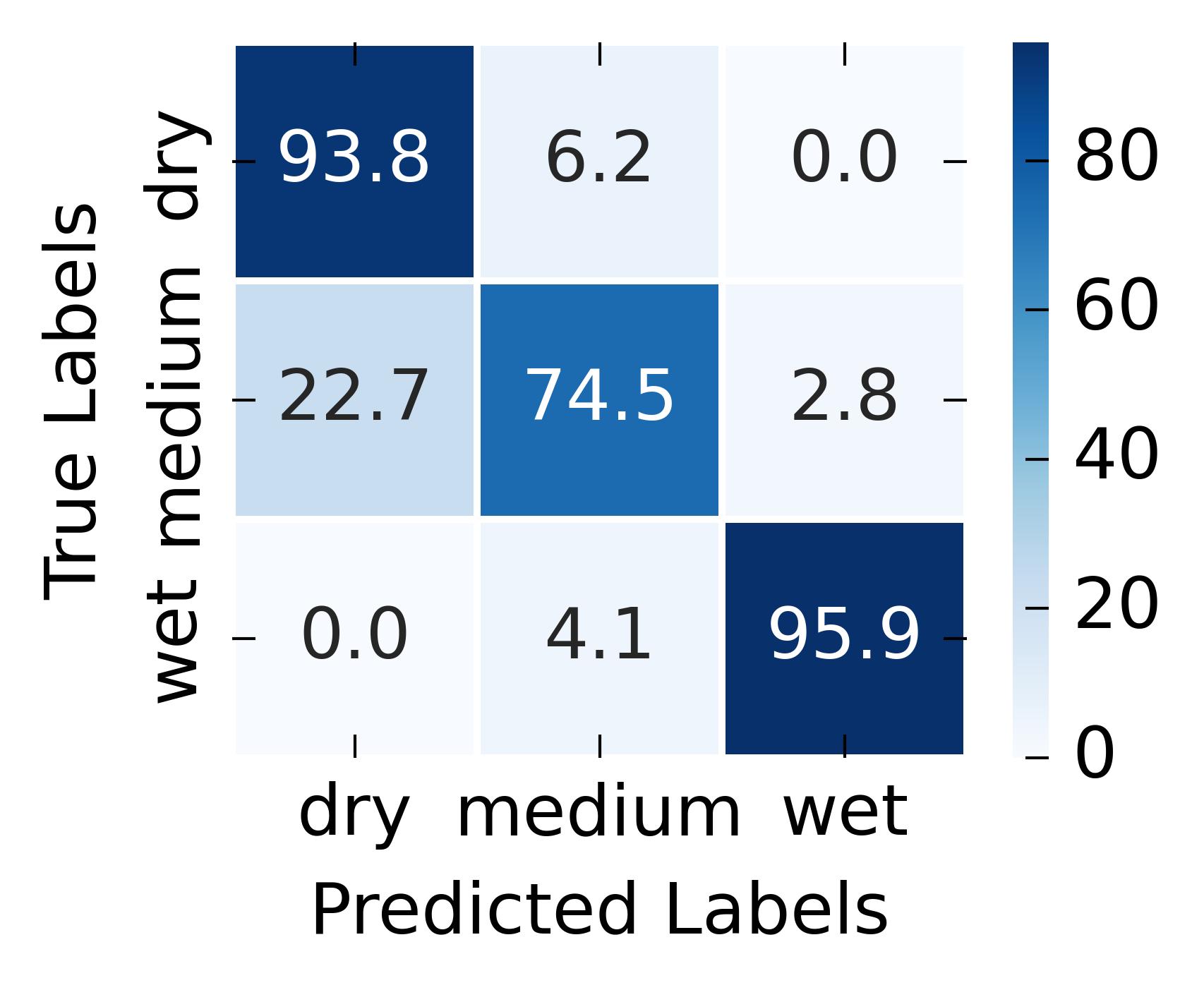}
    \caption{MoistNetLite}
  \end{subfigure}
  \centering
  \begin{subfigure}[b]{0.3\textwidth}
    \centering
    \includegraphics[width=\textwidth]{ 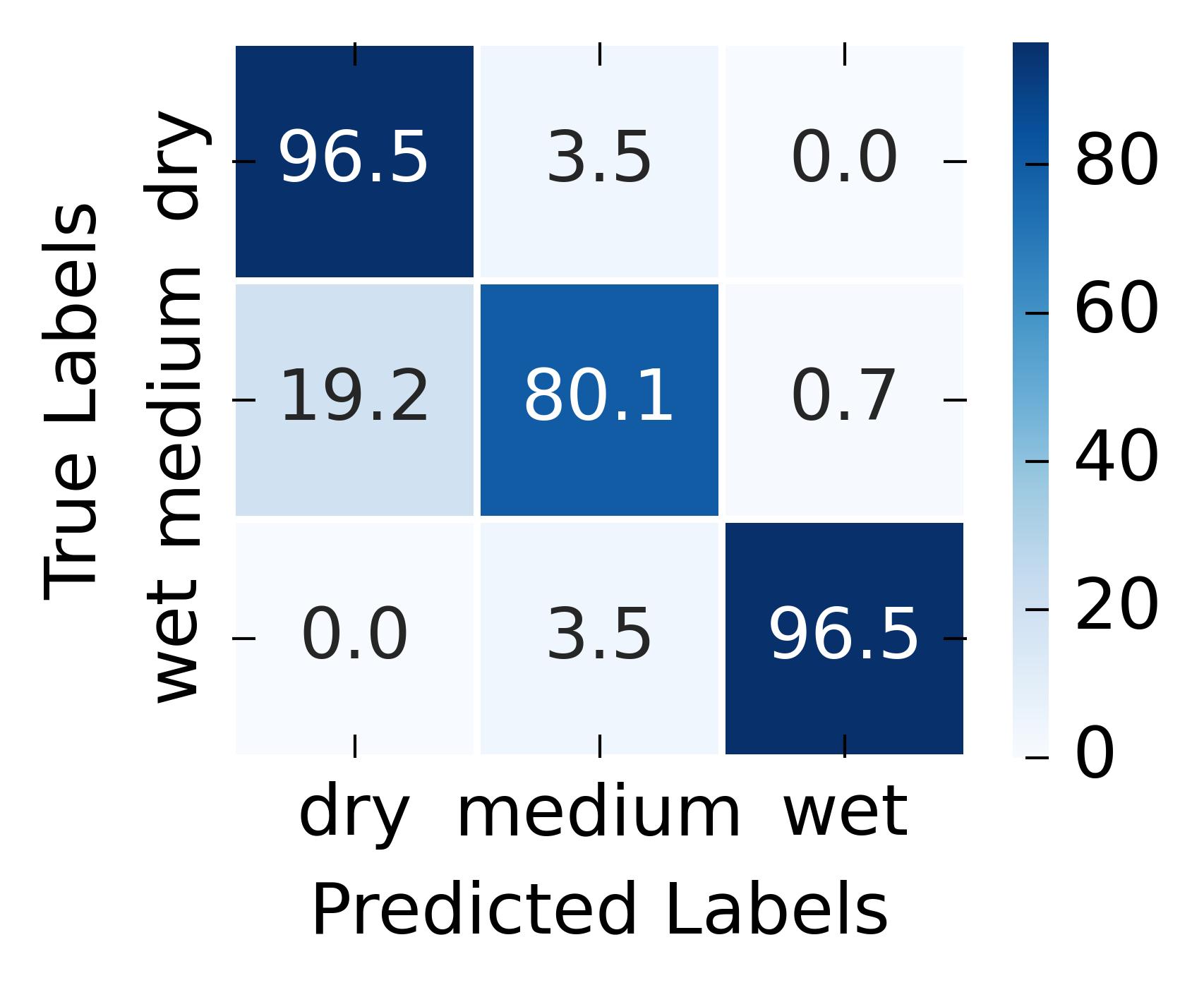}
    \caption{MoistNetMax}
  \end{subfigure}
  \begin{subfigure}[b]{0.3\textwidth}
    \centering
    \includegraphics[width=\textwidth]{ 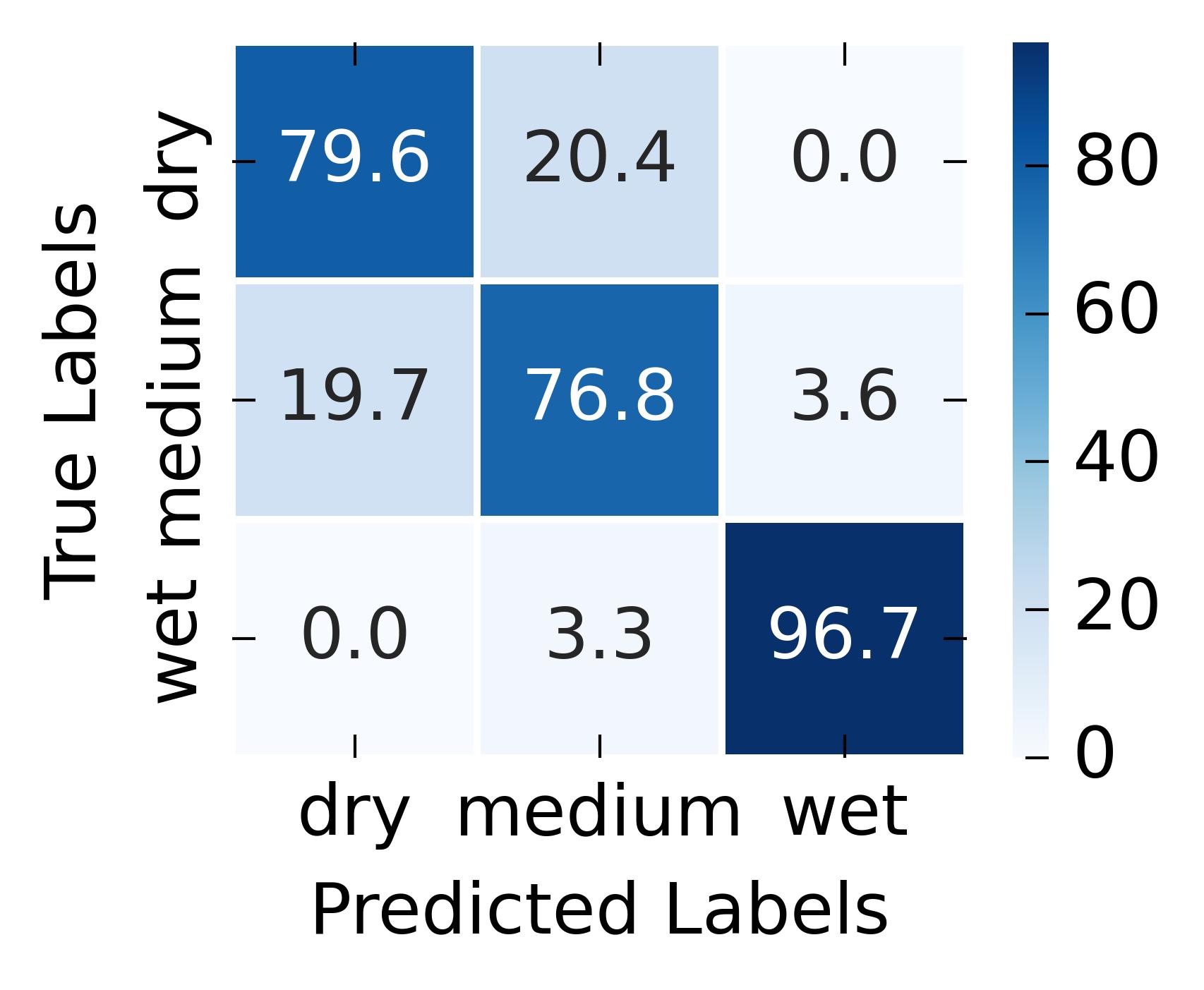}
    \caption{ResNet152V2}
  \end{subfigure}
  \begin{subfigure}[b]{0.3\textwidth}
    \centering
    \includegraphics[width=\textwidth]{ 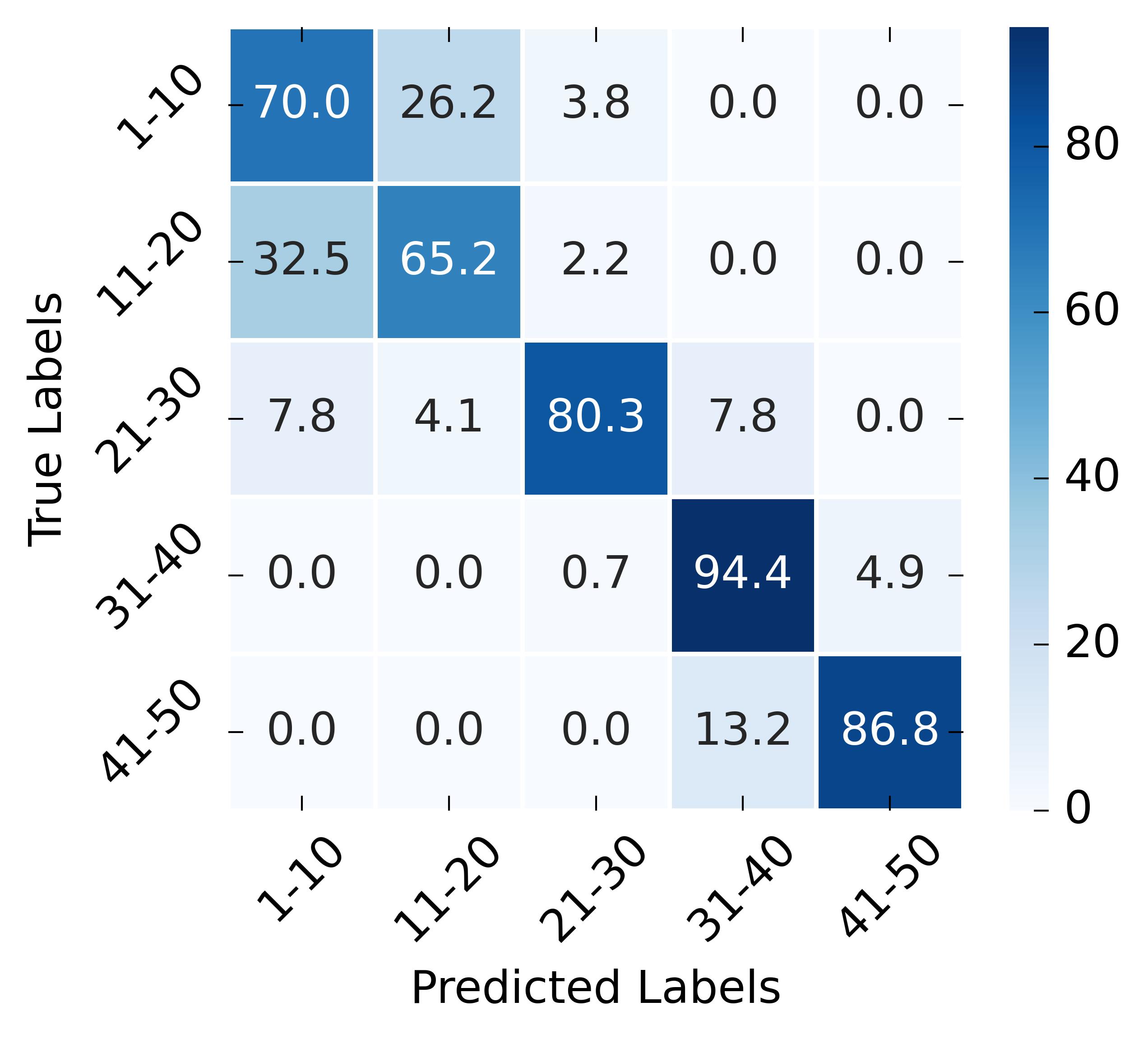}
    \caption{MoistNetLite}
  \end{subfigure}
  \centering
  \begin{subfigure}[b]{0.3\textwidth}
    \centering
    \includegraphics[width=\textwidth]{ 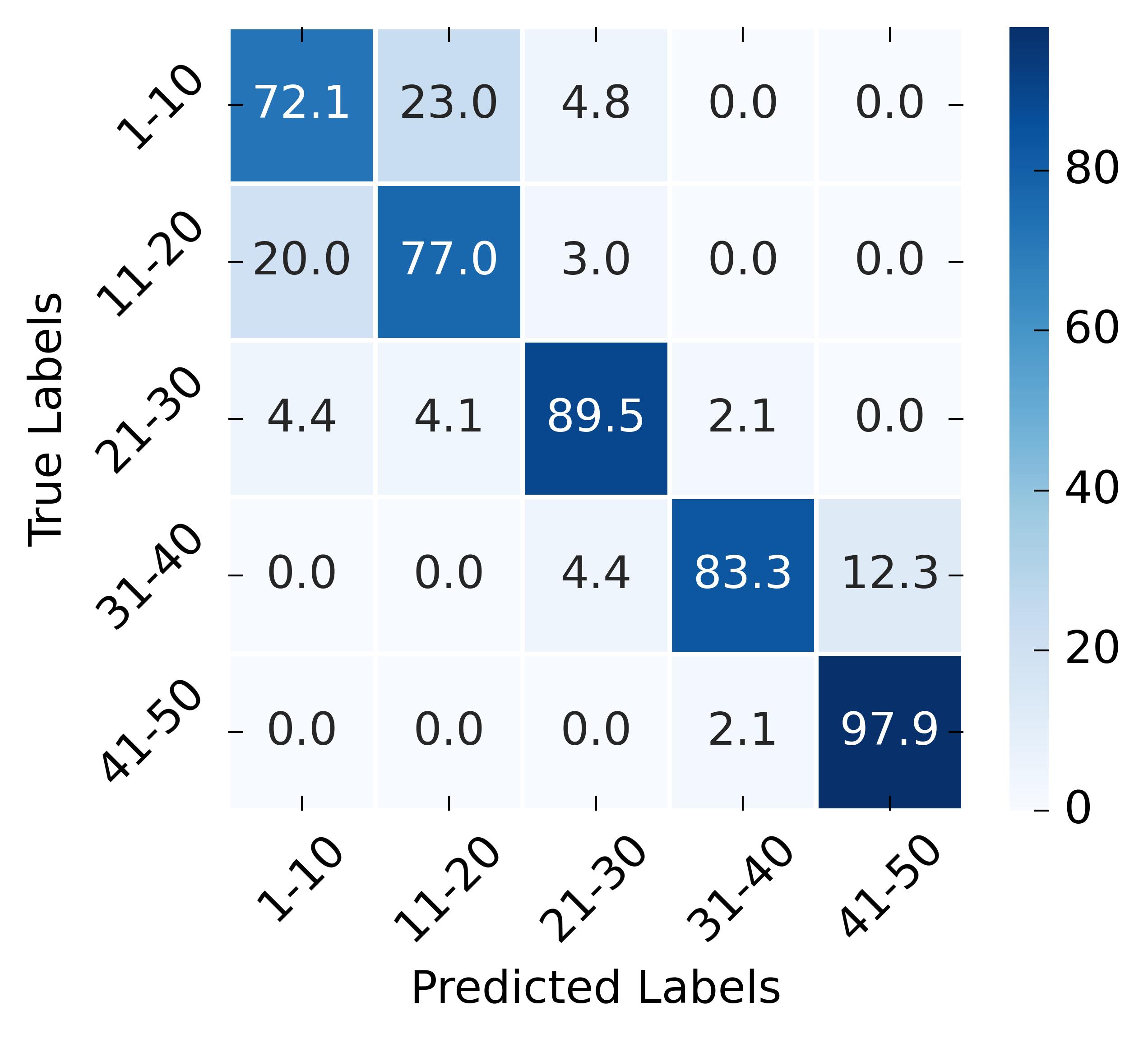}
    \caption{MoistNetMax} 
  \end{subfigure}
  \begin{subfigure}[b]{0.3\textwidth}
    \centering
    \includegraphics[width=\textwidth]{ 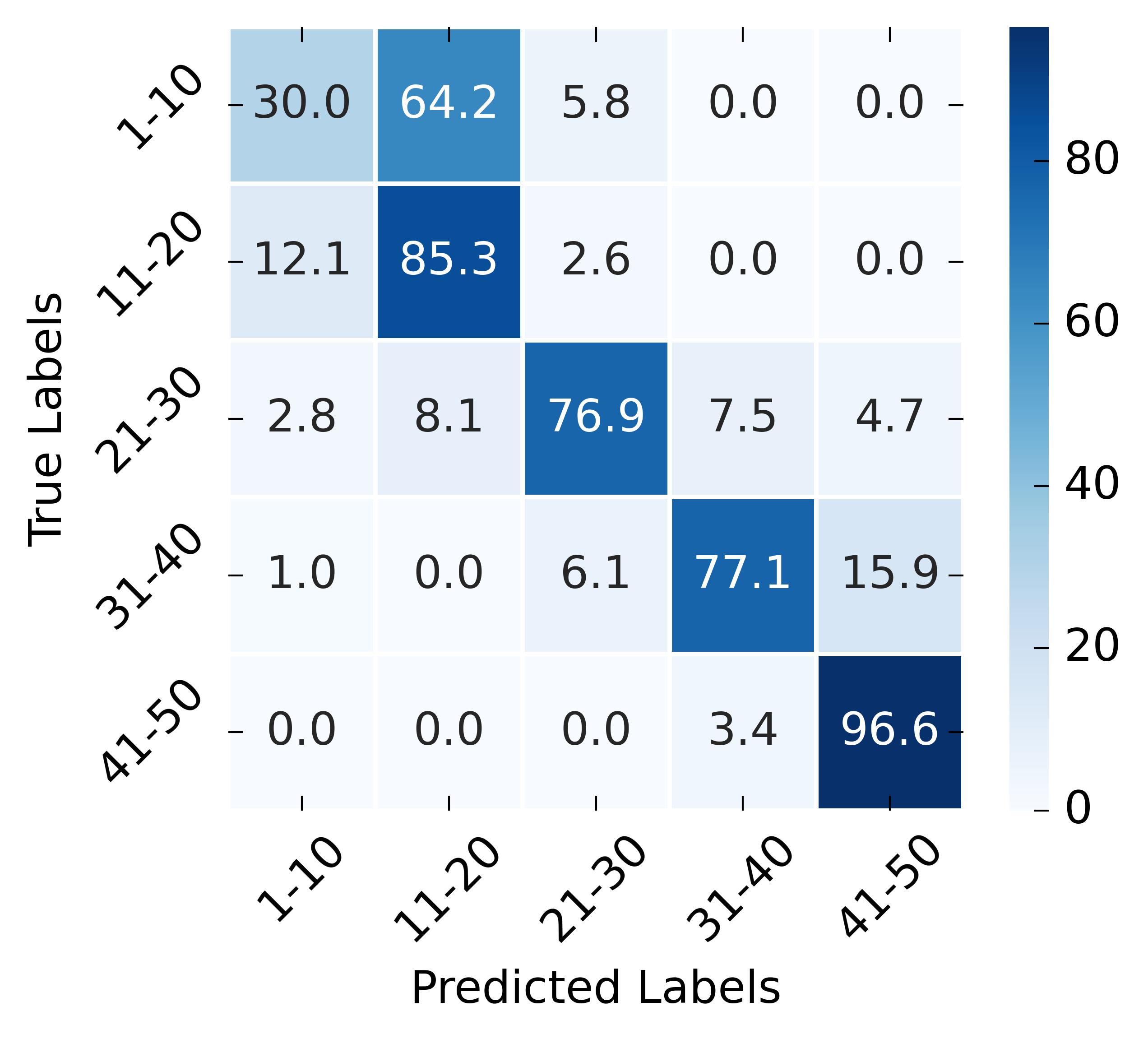}
    \caption{ResNet152V2}
  \end{subfigure}
  \caption{(Best viewed in color) Confusion matrix of the predictions of MoistNet models and the top baseline ResNet152V2. These confusion matrices have been generated with the predictions on the test dataset. Entries of the confusion matrices were converted to percentages to gain better insights. The top row is for three-class, and the bottom row is for five-class classifications.}
  \label{fig:conf_matrix}
\end{figure}

We found some baselines, including VGG \citep{simonyan2014very} and NASNet \citep{zoph2018learning} architectures, which did not converge at all when we trained them from scratch. Among all other baselines, ResNet architectures (especially ResNet50, ResNet152, and ResNet152V2), and InceptionV3 demonstrated commendable performance with 83\% F1-score. ResNet50 and InceptionV3 appeared to have less number of model parameters and showed faster inference speed compared to ResNet152 and ResNet152V2. MoistNetLite showed an increased F1-score of 87\% with a very small number of model parameters (0.29 million) and a very fast inference speed (0.58 ms/image). These exceptional properties of MoiseNetLite make it suitable for applications on resource-constrained devices such as smartphones and single-board computers. 

MoistNetMax, on the other hand, exhibited the most accurate results achieving an F1-score of 91\% with the additional computational load. {\color{myblue}To check whether the performance improvement of MoistNet models is significant or not, we performed the Friedman test with the results of sixteen baseline models across four performance metrics (accuracy, precision, recall, and F1-score). We defined the null hypothesis as follows: there is no significant difference in the performance of the models. For each metric, we ranked the performance scores across all models. If two or more models had the same performance score, we assigned them the average rank. We calculated the Friedman test statistic using Equation \ref{eqn:friedman}. 
\begin{equation}
\label{eqn:friedman}
\chi^2 = \frac{12}{n m (m + 1)} \left( \sum_{j=1}^{n} R_j^2 - \frac{m (m + 1)^2}{4} \right)
\end{equation}
where, $n$ is the number of models compared, $m$ is the number of metrics used, and $R_j$ is the sum of ranks for the $j$-th model across all metrics. Here, the test statistic $\chi^2$ follows a chi-square distribution with $m-$1 degrees of freedom. We found the Friedman test statistic and $p$-value to be 28.35 and $3.05\times 10^{-6}$, respectively. Since the $p$-value is less than 0.05, we reject the null hypothesis and conclude that there are significant differences between the performance of the models. Consequently, we can conclude that the MoistNet models showed improved classification performance in wood chip MC class prediction compared to the baselines.}

{\color{myblue}Figure \ref{fig:conf_matrix}(a)-(c) illustrate the confusion matrix for three class MC predictions of the top three models (MoistNetLite, MoistNetMax, and ResNet152V2). The class \textit{`medium'} appeared to be the most problematic one for the MoistNet architectures. For example, in \ref{fig:conf_matrix}(a) and (b), respectively, 22.7\% and 19.2\% of the samples were predicted as \textit{`dry'} when they were actually \textit{`medium'}. On the contrary, in the case of the baseline ResNet152V2, both the \textit{`dry'} and \textit{`medium'} classes seemed to be the most challenging ones.}

\begin{figure}[tbp]
  \centering
  \begin{subfigure}[b]{0.3\textwidth}
    \centering
    \includegraphics[width=\textwidth]{ 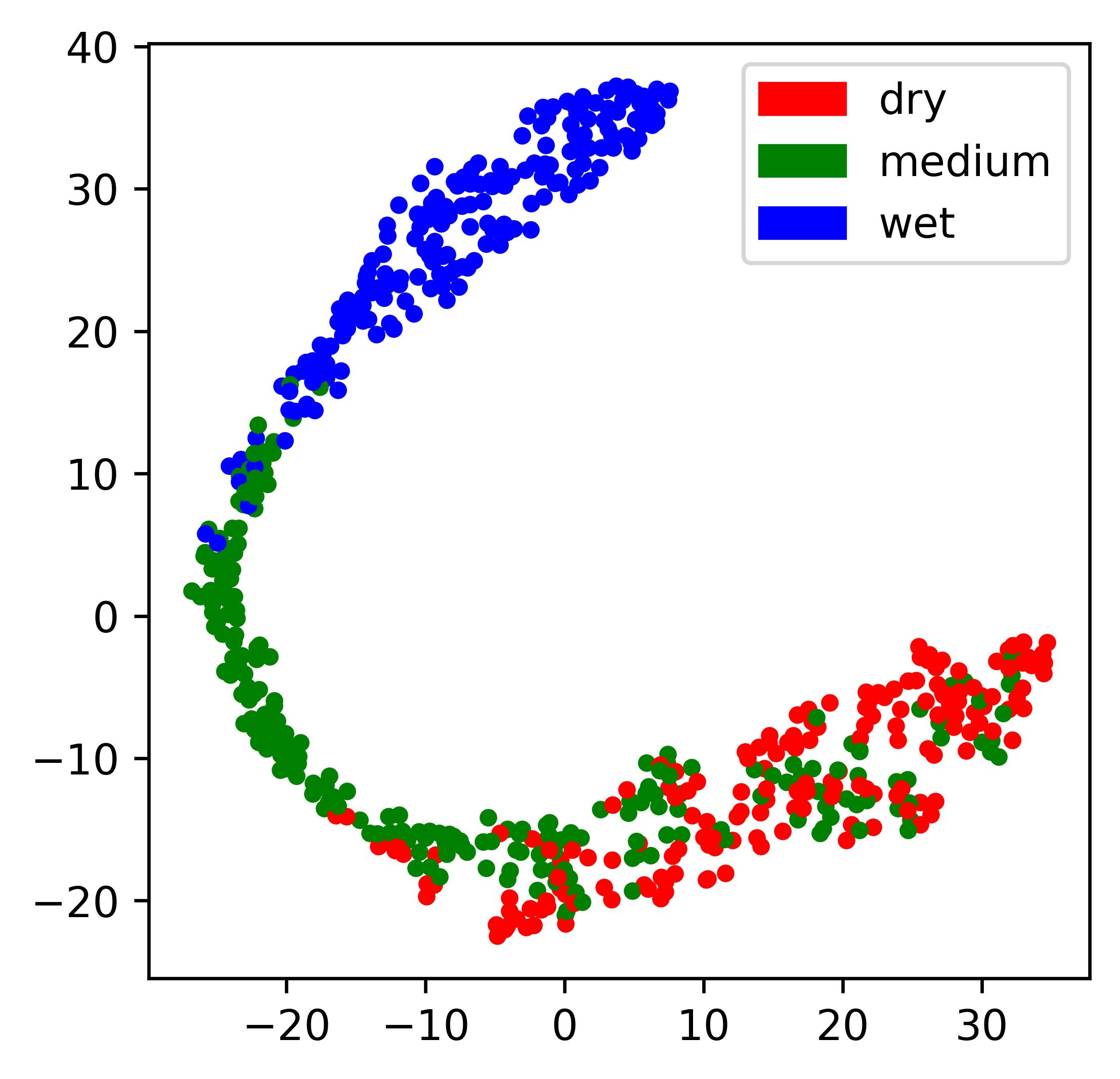}
    \caption{MoistNetLite}
  \end{subfigure}
  \centering
  \begin{subfigure}[b]{0.3\textwidth}
    \centering
    \includegraphics[width=\textwidth]{ 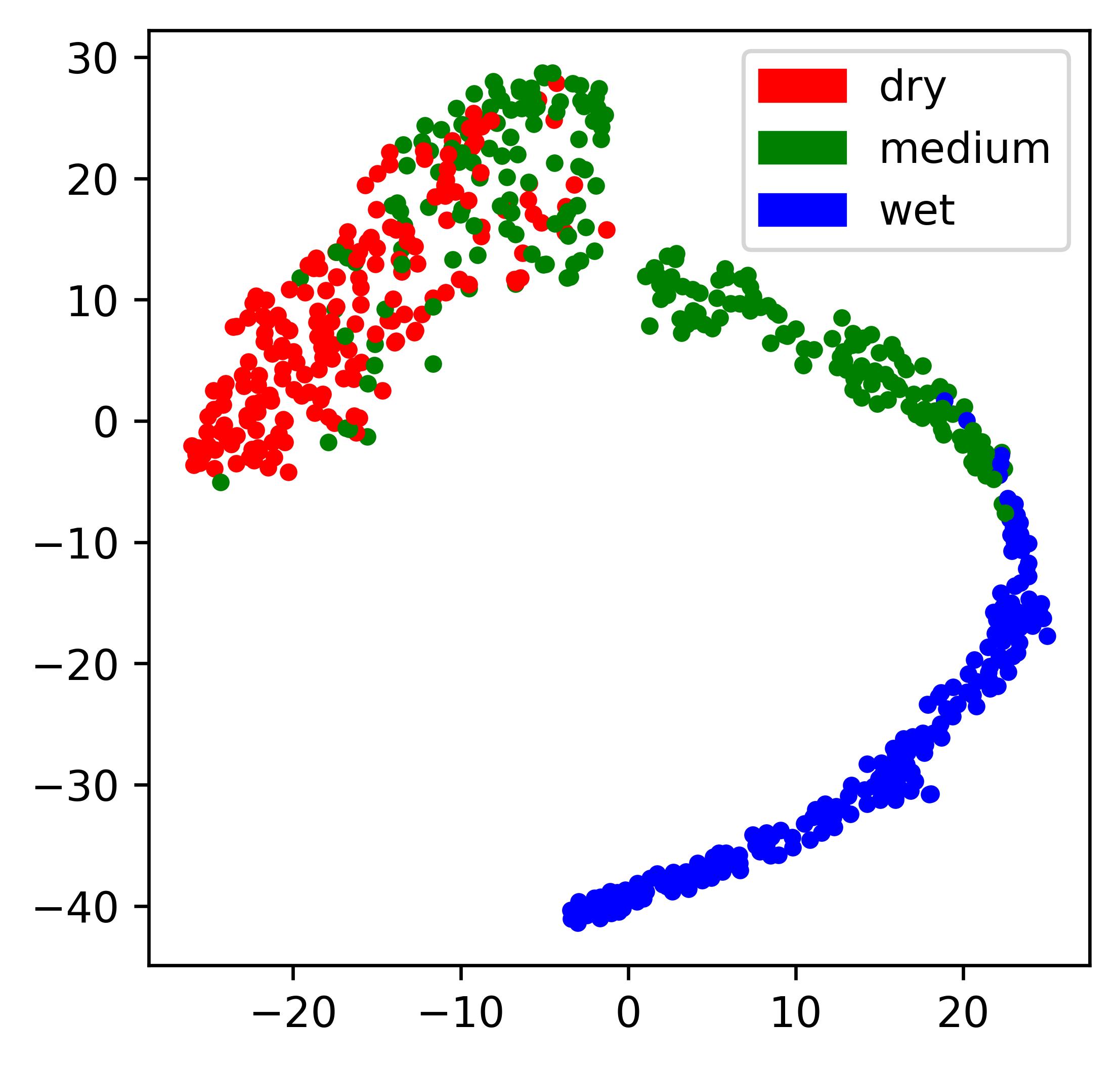}
    \caption{MoistNetMax}
  \end{subfigure}
  \begin{subfigure}[b]{0.3\textwidth}
    \centering
    \includegraphics[width=\textwidth]{ 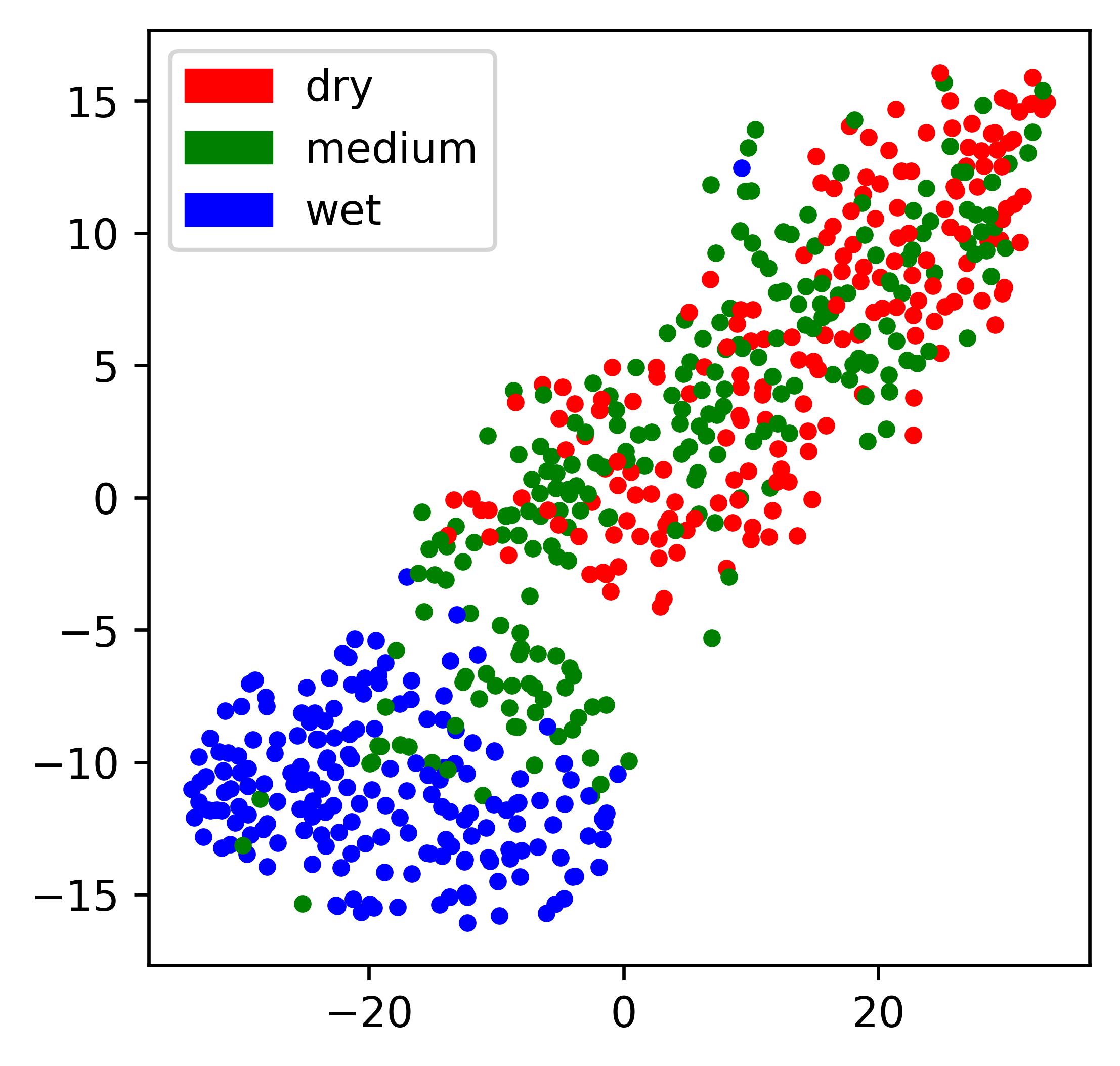}
    \caption{ResNet152V2}
  \end{subfigure}
  \caption{(Best viewed in color) t-SNE plots of the final layer embedding of MoistNet models and the top baseline ResNet152V2.}
  \label{fig:tsne}
\end{figure} 

We have also explored the learned embedding of these three models to understand how well the models could classify the MC. To complete that, t-SNE plots \citep{wattenberg2016use} were generated from the final layer embedding of the models. Figure \ref{fig:tsne} illustrates the t-SNE plots for three class classifications. As it appears, MoistNetLite and MoistNetMax generated easily distinguishable classes, whereas, in the case of the ResNet152V2 model, the classes were more cluttered. {\color{myblue}Another notable fact was that the pair of \textit{`dry'} and \textit{`medium'} classes were closer and less distinguishable compared to the pair of \textit{`medium'} and \textit{`wet'} classes which could also be supported by the confusion matrices in Figure \ref{fig:conf_matrix}(a)-(c).

Next, we explored how the performance of the models changes if we increase the number of classes. For example, we considered five classes: \textit{1-10\%, 11-21\%, 21-30\%, 31-40\%}, and \textit{41-50\%} and performed similar experiments. Table \ref{tab:class} shows the performance comparison of the models for three and five classes. we observed a 6.8\%, 6.5\%, and 3.6\% drop in F1-score for the MoistNetLite, MoistNetMax, and ResNet152V2, respectively. Figure \ref{fig:conf_matrix}(d)-(f) demonstrates the confusion matrix for five class classifications of the MC. As a matter of fact, \textit{1-10\%} and \textit{11-21\%} classes were the most challenging classes to distinguish for all models. From the confusion matrix of both three-class and five-class classifications, it could be inferred that \textit{dry} samples are less distinguishable than \textit{wet} samples.}
\begin{table}[htbp]
\centering
\caption{Moisture content class prediction performance considering three and five classes on the dataset from source 1.}
\label{tab:class}
\resizebox{\textwidth}{!}{%
\begin{tabular}{@{}ccccccc@{}}
\hline
\# Classes   & \multicolumn{3}{c}{Three}     & \multicolumn{3}{c}{Five}       \\ \hline
Mertrics     & Precision & Recall & F1-score & Precision & Recall & F1-score \\ \hline
ResNet152V2  &  0.84 ± 0.04        &   0.83 ± 0.04     &   0.83 ± 0.04       &    0.81 ± 0.01 & 0.80 ± 0.00 & 0.80 ± 0.01       \\
MoistNetLite &   0.87 ± 0.05        &   0.87 ± 0.05     &      0.87 ± 0.05    &   0.82 ± 0.02        &    0.80 ± 0.03    &     0.81 ± 0.02  \\
MoistNetMax  &     \textbf{0.91 ± 0.02}      &    \textbf{0.91 ± 0.02}   &      \textbf{0.91 ± 0.02}     &     \textbf{0.85 ± 0.02}    &   \textbf{0.85 ± 0.02}   & \textbf{0.85 ± 0.02} \\ \hline
\end{tabular}%
}
\end{table}

\subsection{Sensitivity Analysis}
\subsubsection{Change in Dataset}
Until now all analysis was carried out on the dataset from source 1. However, we obtained wood chips from an alternative source (referred to as `source 2') in two distinct batches labeled as `batch 1' and `batch 2'. Consequently, we conducted analogous experiments mentioned previously on batch 1, and batch 2, as well as the complete set of images from source 2, in order to assess the resilience of the proposed models. Table \ref{tab:source2} demonstrates the performance of MoistNet models and the baseline model ResNet152V2. Results indicate the superior performance of the MoistNet models over the baseline across every performance metric. Although the overall classification performance remained similar when the dataset was changed from source 1 to source 2, a drop in the performance could be observed in the case of source 2 - batch 2. 
\begin{table}[htbp]
\centering
\caption{Moisture content class prediction performance on the dataset from source 2. }
\label{tab:source2}
\resizebox{\textwidth}{!}{%
\begin{tabular}{@{}cccccccccc@{}}
\hline
Dataset  & \multicolumn{3}{c}{Source 2 - Batch 1} & \multicolumn{3}{c}{Source 2 - Batch 2} & \multicolumn{3}{c}{Entire Source 2} \\ \hline
Mertrics & Precision    & Recall    & F1-score    & Precision    & Recall    & F1-score    & Precision   & Recall   & F1-score  \\ \hline
ResNet152V2  & 0.83 ± 0.04 & 0.83 ± 0.04 & 0.83 ± 0.04 & 0.73 ± 0.05 & 0.69 ± 0.04 & 0.71 ± 0.03 & 0.82 ± 0.03 & 0.82 ± 0.03 & 0.82 ± 0.03 \\
MoistNetLite & 0.86 ± 0.02 & 0.86 ± 0.02 & 0.86 ± 0.02 & 0.83 ± 0.02 & 0.83 ± 0.02 & 0.83 ± 0.02 & 0.86 ± 0.03
 & 0.86 ± 0.03
 & 0.86 ± 0.03
 \\
MoistNetMax  & \textbf{0.90 ± 0.01} & \textbf{0.90 ± 0.01} & \textbf{0.90 ± 0.01} & \textbf{0.84 ± 0.03} & \textbf{0.84 ± 0.03} & \textbf{0.84 ± 0.03} & \textbf{0.90 ± 0.01}
 & \textbf{0.90 ± 0.01}
 & \textbf{0.90 ± 0.01}
 \\ \hline
\end{tabular}%
}
\end{table}
\begin{figure}[htbp]
  \centering
  \begin{subfigure}[b]{0.3\textwidth}
    \centering
    \includegraphics[width=\textwidth]{ 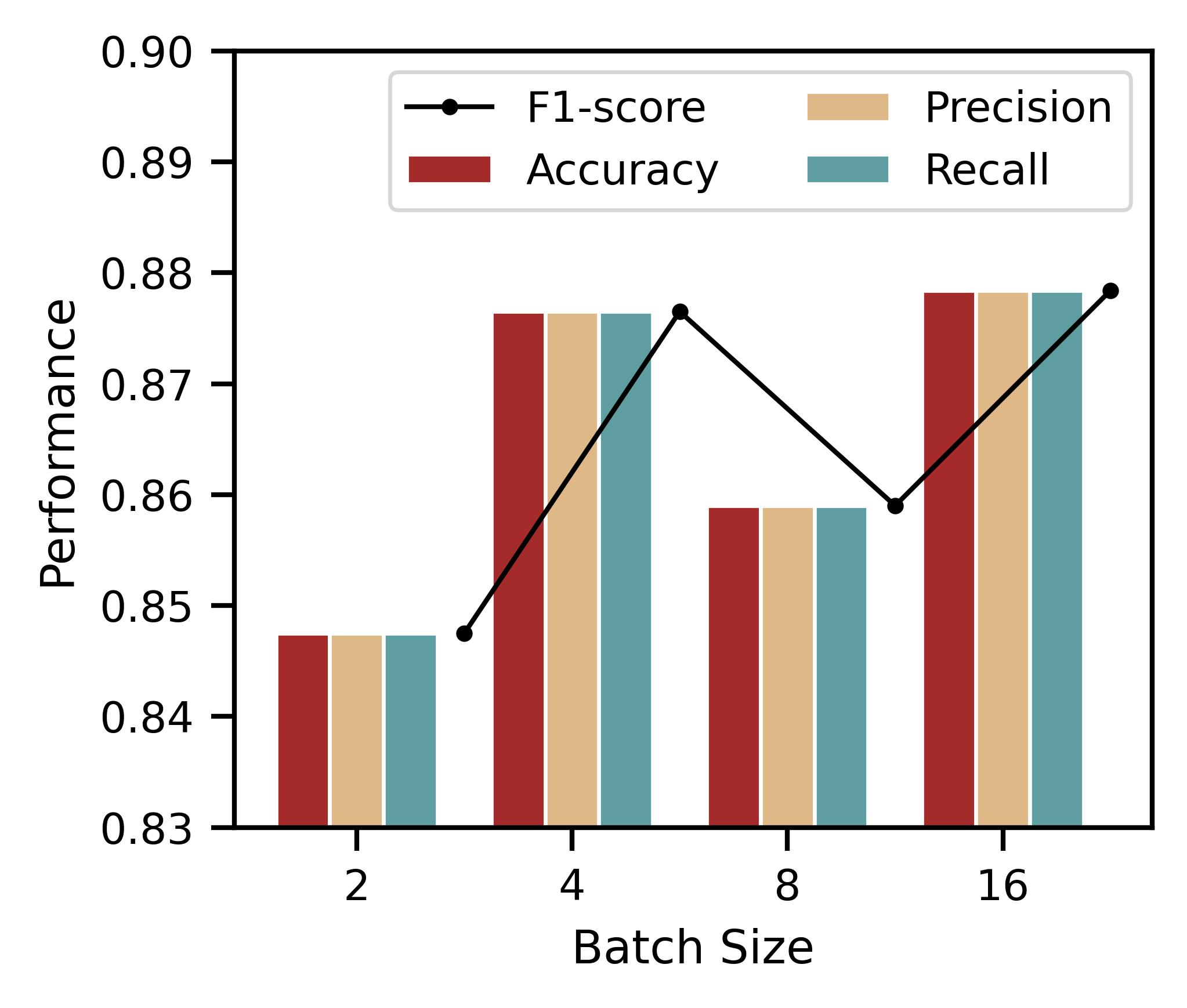}
    \caption{MoistNetLite with different batch sizes}
  \end{subfigure}
  \centering
  \begin{subfigure}[b]{0.3\textwidth}
    \centering
    \includegraphics[width=\textwidth]{ 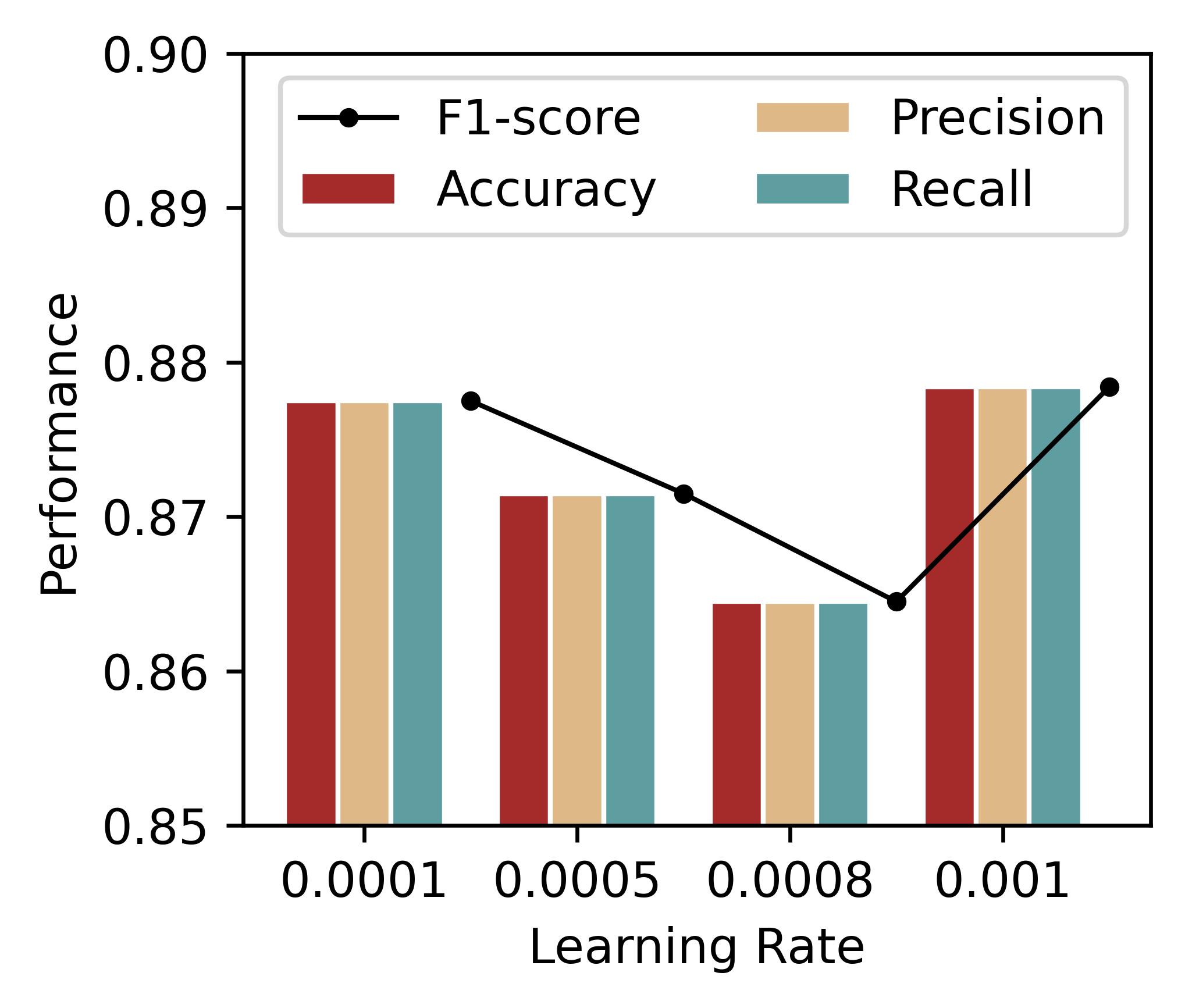}
    \caption{MoistNetLite with different learning rates}
  \end{subfigure}
  \begin{subfigure}[b]{0.3\textwidth}
    \centering
    \includegraphics[width=\textwidth]{ 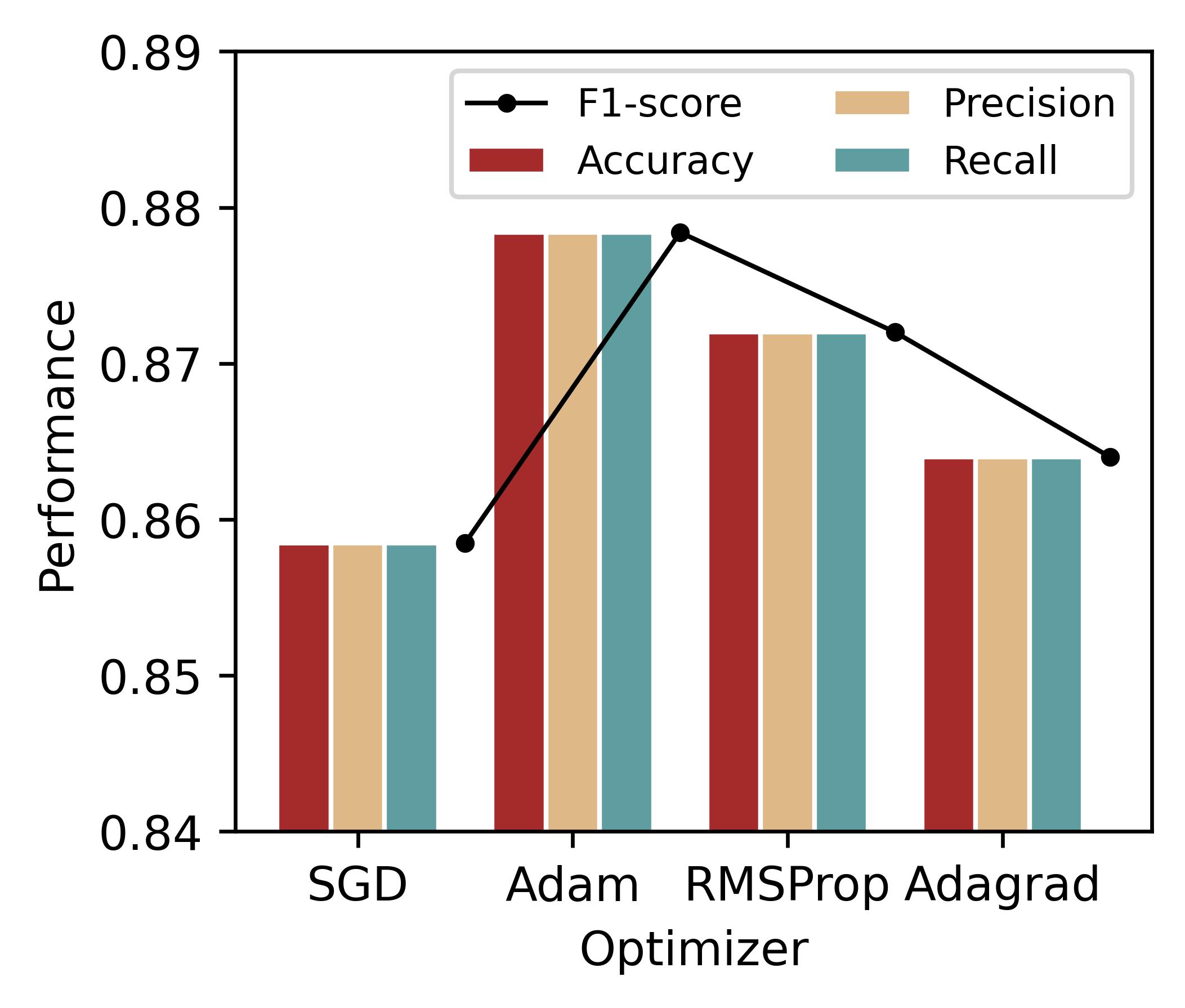}
    \caption{MoistNetLite with different optimizers}
  \end{subfigure}
    \begin{subfigure}[b]{0.3\textwidth}
    \centering
    \includegraphics[width=\textwidth]{ 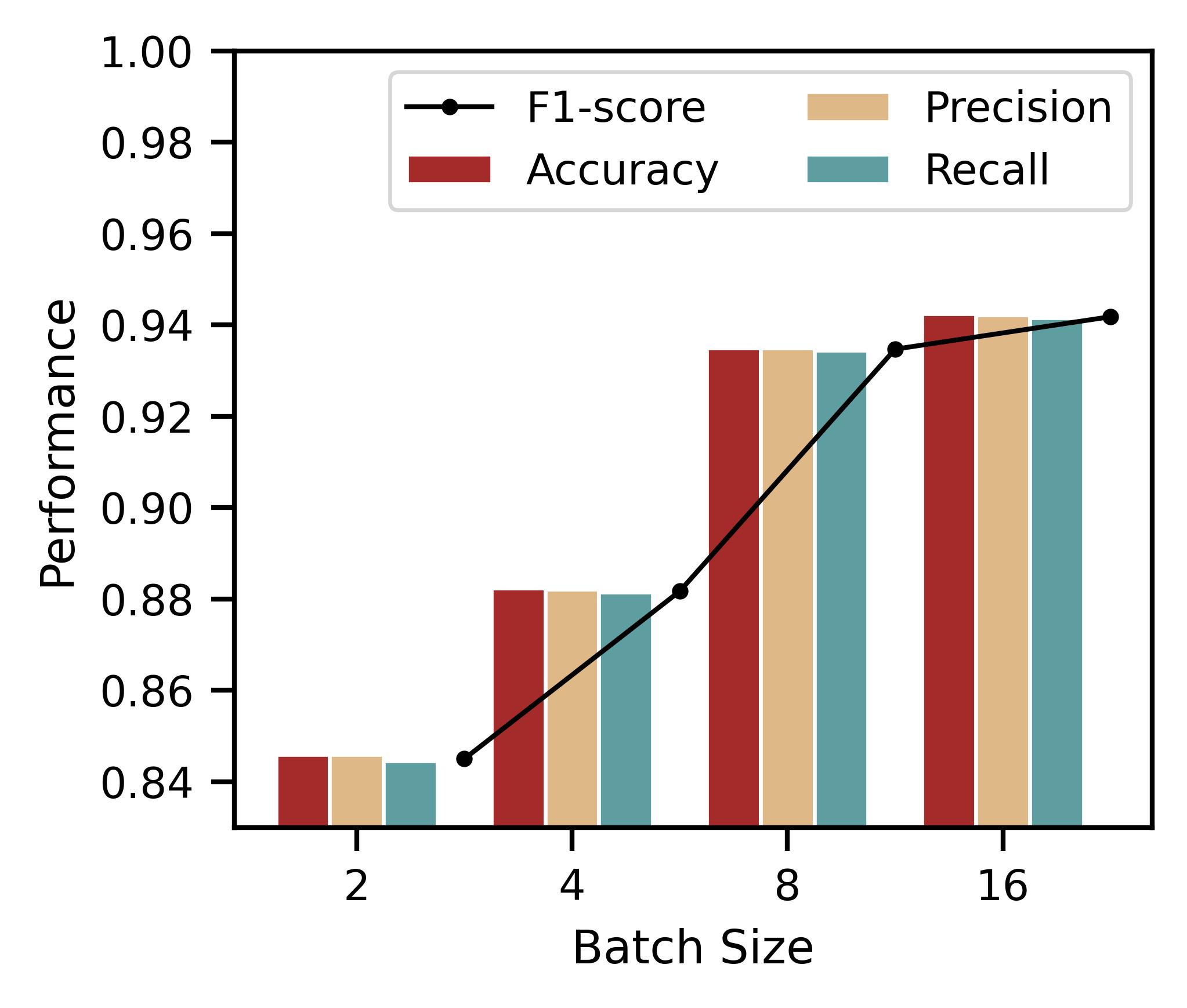}
    \caption{MoistNetMax with different batch sizes}
  \end{subfigure}
  \centering
  \begin{subfigure}[b]{0.3\textwidth}
    \centering
    \includegraphics[width=\textwidth]{ 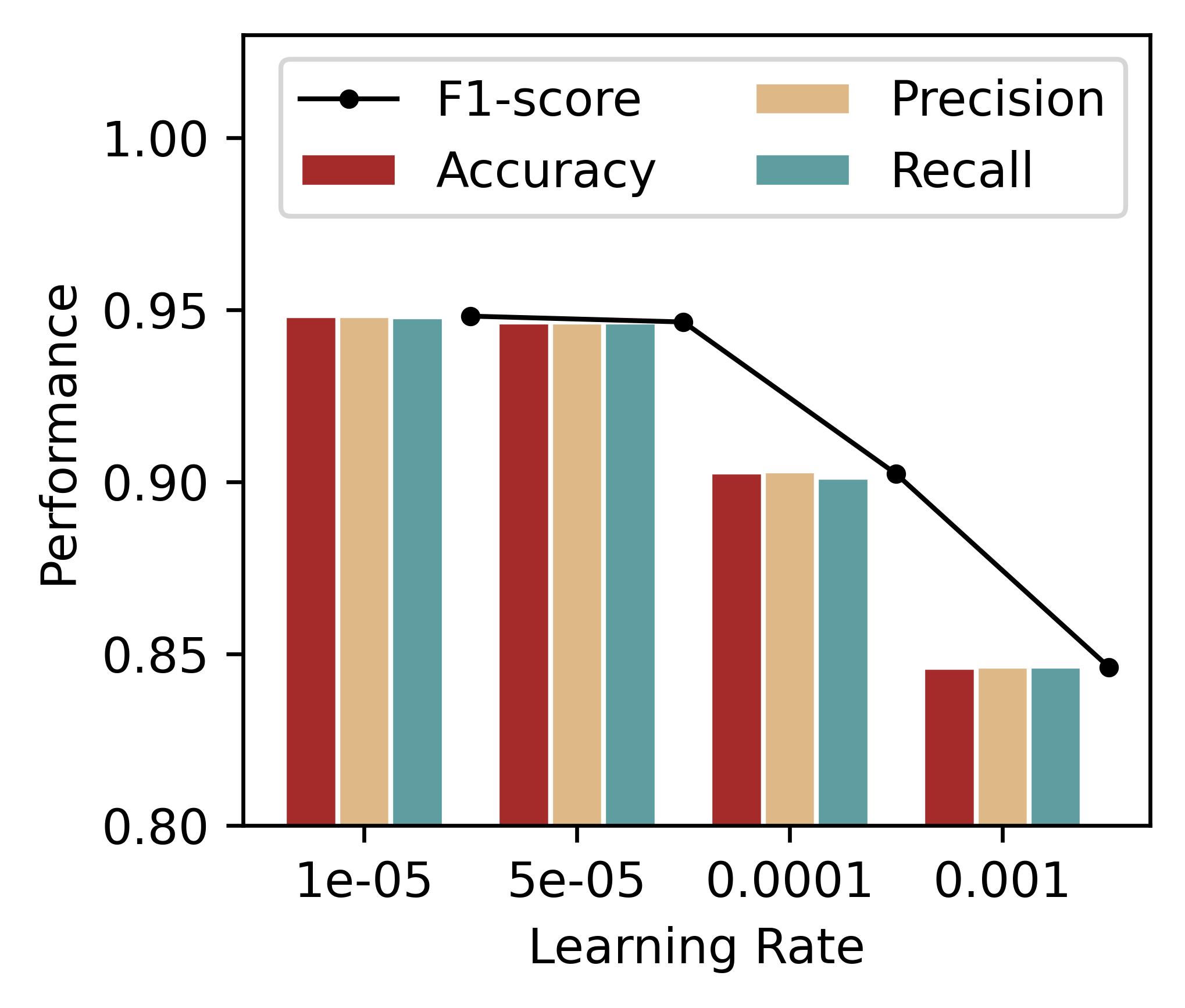}
    \caption{MoistNetMax with different learning rates}
  \end{subfigure}
  \begin{subfigure}[b]{0.3\textwidth}
    \centering
    \includegraphics[width=\textwidth]{ 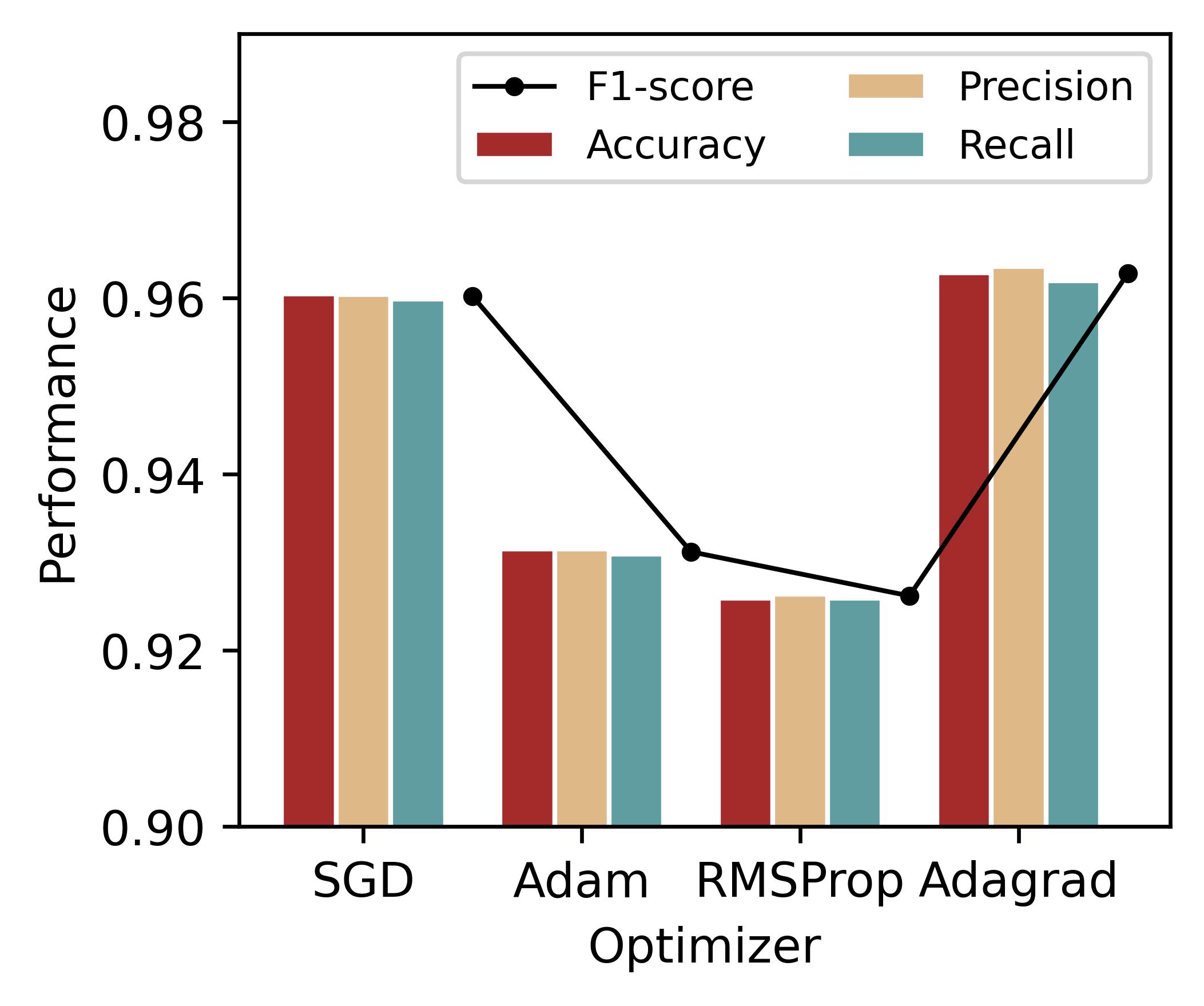}
    \caption{MoistNetMax with different optimizers}
  \end{subfigure}
    \begin{subfigure}[b]{0.3\textwidth}
    \centering
    \includegraphics[width=\textwidth]{ 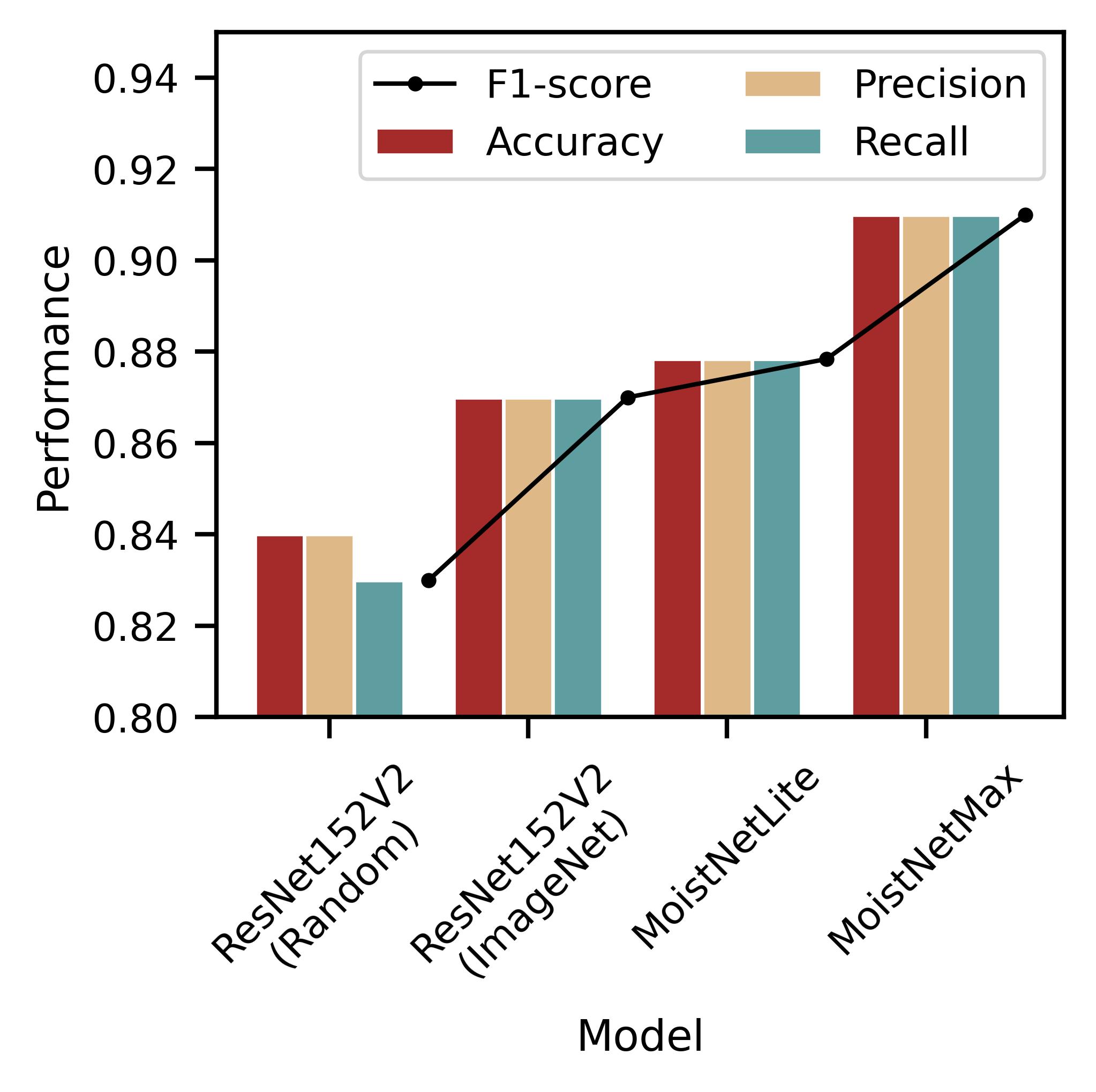}
    \caption{Different weight initialization methods}
  \end{subfigure}
  \caption{(Best viewed in color) Sensitivity analysis of the MoistNetLite and MoistNetMax models in terms of batch size, learning rate, optimizer, and weight initialization.}
  \label{fig:sensitivity}
\end{figure}

\subsubsection{Change in Cross-Validation Setting}
\label{sec:cross}
In our study, we implemented a K-fold cross-validation strategy with $K = 4$. It is common practice in machine learning modeling to utilize 5-fold or 10-fold cross-validation. However, we specifically chose $K = 4$ for a particular reason. Our dataset consisted of image data from various classes, such as wood chips with different MCs, arranged in four trays. To ensure a comprehensive evaluation, we captured 20 images from each tray by shuffling the wood chips before image capture. We aimed to perform a group K-fold cross-validation, treating images taken from the same tray as a group. As there were four trays for each class, we set the value of K as 4 for both types of cross-validation. To compare the performance of the 18 models, as shown in Table \ref{tab:ttest} (see \ref{sec:appendix_c}), between K-fold and Group-K-fold cross-validation, we conducted a paired t-test on the F1-scores. The purpose was to determine if there were any significant differences in performance between the two cross-validation approaches. We formulated the null hypothesis and alternate hypothesis as: \\
$H_0$: There is no significant difference between the K-fold and Group-K-fold results.\\
$H_1$: There is a significant difference between the K-fold and Group-K-fold results.\\

We calculated the value of the test statistic $t$ using Equation \ref{eqn:ttest}:
\begin{equation}
    \label{eqn:ttest}
    t = \frac{\bar{x}_{diff}}{\frac{s_{diff}}{\sqrt{n}}}
\end{equation}
where $\bar{x}_{diff}$ indicates the sample mean of the differences between F1-scores using two types of cross-validation, $s_{diff}$ means the standard deviation of the same differences, and $n$ is the number of models used in this study. We got a $t$ value of 0.3878. Using this $t$ and degrees of freedom ($df$) equal to 17, we found the $p$-value to be 0.7029. Since the $p$-value is greater than our significance level of $\alpha = 0.05$, we failed to reject the null hypothesis $H_0$. This implies that there is no significant difference between the results obtained from K-fold and Group-K-fold cross-validation. Based on this analysis, we can conclude that utilizing trays to create wood chip samples does not have a significant impact on the experimental results.

\subsubsection{Change in Hyperparameters}
{\color{myblue}To assess the influence of hyperparameters, including batch size, learning rate, optimizer, and weight initialization, on the performance of MoistNet models, a sensitivity analysis was conducted. To expedite the analysis, the experiments were carried out on a single fold instead of utilizing cross-validation. 

In the case of batch size, it was observed that both MoistNet models displayed sensitivity to this hyperparameter. Figure \ref{fig:sensitivity}(a) and (d) illustrate that the highest accuracy was achieved when the batch size was set to 16 for both models. Notably, for MoistNetLite, a batch size of 4 also yielded comparable results. It is important to mention that we could not explore batch sizes of more than 16 due to computational resource constraints.

While changing the learning rate, we found both models to be sensitive. Performance of the MoistNetLite showed a downward trend with an increase in the learning rate except for 0.001, as shown in Figure \ref{fig:sensitivity}(b). MoistNetLite achieved the highest precision at a 0.001 learning rate. However, a learning rate of more than 0.001 did not result in convergence. MoistNetMax, on the other hand, showed a consistent increase in performance when the learning rate was decreased, as shown in \ref{fig:sensitivity}(e), and achieved the best result with a learning rate of $10^{-5}$. We did not try learning rate smaller than $10^{-5}$ because the performance started to flatten from $5\times10^{-5}$ to $10^{-5}$.

Optimizers play one of the most critical roles in deep learning model training. In this study, we found Adam to be the best-performing optimizer for the MoistNetLite model, as shown in Figure \ref{fig:sensitivity} (c). On the other hand, Figure \ref{fig:sensitivity} (f) shows that Adagrad and SGD performed better for MoistNetMax. Weight initialization is another important parameter to consider. There are several ways to initialize the weights of the model, including random initialization, starting with pre-trained imagenet weights and starting with weights that are trained on the same wood chip data. In this experiment, we compared the first two of the mentioned initialization methods. Figure \ref{fig:sensitivity} (g) demonstrates that the performance of the ResNet152V2 has improved when the imagenet weights were used in place of random weights. However, the MoistNetLite and MoistNetMax still performed better than the improved version of ResNet152V2. }

{\color{myblue}
\subsection{Broader Implications of MoistNet}
While the current MC determination method requires hours to provide precise measurements, the proposed image-based MoistNet method gets it done instantaneously. Thus, it is possible to optimize the drying time by efficiently assessing the moisture content of incoming woodchips to achieve the desired moisture level. This efficient and precise way of tuning the drying process ensures that the woodchips are not excessively dried, which not only conserves energy but also reduces emissions and unnecessary fuel costs. Additionally, the MC of woodchips is critical in pelletizing, affecting the binding characteristics of the pellets, their durability, and energy content. 

In addition to the wood pellet industry, the MC of wood chips has a huge impact on the pulp and paper industry. The pulping behavior and the energy requirements for pulping and drying processes are largely dependent on the MC measurement. Thus, a fast and precise assessment of MC can lead to energy savings and increased production efficiency. In other wood chip-reliant fuel production facilities, knowing the moisture content allows for better control of combustion processes, reducing emissions of pollutants and helping facilities comply with environmental regulations. Essentially, this MoistNet-based approach promotes efficient drying by preventing over-drying, thereby minimizing environmental impact and operational expenses. }

\section{Conclusions and Future Work}
{\color{myblue}Moisture content measurement is a crucial and time-consuming task in the wood chip-reliant industries. To overcome the shortcomings of the existing methods, we have proposed an image-based solution to the wood chip MC class prediction task. First, an extensive image dataset with carefully curated labels has been developed. Then we proposed two deep learning model architectures, namely MoistNetLite and MoistNetMax, generated through NAS and hyperparameter optimization. MoistNetLite is a lightweight and fast model capable of achieving state-of-the-art performance with substantially reduced inference time. On the other hand, MoistNetMax improved the prediction performance further with additional computational overhead. The performance of these proposed models has been compared with 16 state-of-the-art deep learning models. Finally, we performed a comprehensive sensitivity analysis that shed light on some key insights.} From this study, we draw the following conclusions:

\begin{itemize}
    \item Machine vision has high potential in wood chip MC class prediction. As the results demonstrate, MoistNetMax could provide instantaneous prediction with an accuracy of over 90\%. Such insights would definitely help industries to make informed decisions in process optimization.
    \item The combination of neural architecture search and hyperparameter optimization could lead to superior deep learning architectures focused on specific tasks.
    \item Since we have an ultimate goal of employing the MoistNet models in mobile devices such as smartphones or any hand-held devices, the lightweight MoistNetLite model could be a suitable option. However, those who prioritize accuracy over inference speed can integrate the MoistNetMax into such mobile devices.  
    \item During the data collection process, the usage of trays doesn't have any significant impact on the performance of the prediction models. 
    \item While developing the multi-class MC prediction model, increasing the number of classes from three to five and eventually decreasing the class moisture ranges resulted in a small drop in accuracy (5.7\% for MoistNetLite and 6.5\% for MoistNetMax).
    \item While trained and evaluated on datasets from a new source (source 2), the prediction accuracy remained within a similar range for both of the MoistNet models. This indicates that the model generated by the NAS and hyperparameter optimization based on one source of wood chips (source 1) can also work on wood chips from another source (source 2). 
    
\end{itemize}

{\color{myblue}There are a couple of future directions that we would like to pursue next. First, the image dataset development process relies on the oven drying process, which is still time-consuming. {\color{myblue}To address this, synthetic image generation through Generative Adversarial Networks (GAN) \citep{goodfellow2020generative} can be a possible solution. In GAN, a generator and a discriminator operate in an adversarial manner to create synthetic images similar to real images. Furthermore, conditional GAN (cGAN) \cite{mirza2014conditional} has the ability to generate images with the desired characteristics when specific moisture content class names are provided as input. However, it is highly important to ensure the validity and authenticity of the generated data before using them. Second, in this study, we have considered MC prediction as a classification task that can predict MC in certain ranges. In the future, we plan to consider it as a regression task with a comprehensive image dataset. We plan to do so by gradually decreasing the interval of the moisture classes.} Third, we evaluated each model by training on the datasets from each source separately. In this study, we did not consider training MoistNet models on images from one source and evaluating them on another. This would lead to the challenge of domain shift, which we plan to address by using transfer learning \citep{pan2009survey} and domain adaptation techniques \citep{ganin2015unsupervised} in the future. {\color{myblue}Finally, the ultimate goal is to implement the trained models into resource-constrained devices to make them applicable on an industrial scale. The MoistNetLite (lightweight model) can be a potential candidate for the task. Initially, we plan to deploy it on a single-board computer (SBC) with an industrial camera connected to the SBC. However, the ultimate aim is to make it applicable to smartphones.}

\section*{CRediT Author Statement}
\textbf{Abdur Rahman:} Conceptualization, Methodology, Software, Formal analysis, Data Curation, Writing - Original Draft, Visualization. \textbf{Jason Street:} Conceptualization, Data Curation, Writing - Review \& Editing, Funding acquisition. \textbf{James Wooten:} Conceptualization, Data Curation, Writing - Review \& Editing, Funding acquisition. \textbf{Mohammad Marufuzzaman:} Conceptualization, Writing - Review \& Editing, Supervision, Funding acquisition. \textbf{Veera G. Gude:} Conceptualization, Funding acquisition. \textbf{Randy Buchanan:} Conceptualization, Funding acquisition. \textbf{Haifeng Wang:} Conceptualization, Methodology, Writing - Review \& Editing, Funding acquisition.

\section*{Declaration of Competing Interest}
The authors declare that they have no known competing financial interests or personal relationships that could have appeared to influence the work reported in this paper.

\section*{Data Availability}
Data will be made available on reasonable request.

\section*{Acknowledgements}
This research was supported by USDA-NIFA (Grant No. 2022-67022-37861 and 2020-67019-30772). The authors gratefully acknowledge this funding support.
\bibliographystyle{elsarticle-harv} 
\bibliography{cas-refs}






\appendix

\section{Search Space $\Phi _1$ for NAS}
\label{sec:appendix_a}

\begin{lstlisting}[style=mystyle]

- normalize (Boolean)
- augment (Boolean)
    - translation_factor (Choice: [0.0, 0.1])
    - horizontal_flip (Boolean)
    - vertical_flip (Boolean)
    - rotation_factor (Choice: [0.0, 0.1])
    - zoom_factor (Choice: [0.0, 0.1])
    - contrast_factor (Choice: [0.0, 0.1])
- block_type (Choice: ['resnet', 'xception', 'vanilla', 'efficient'])
    If block_type == 'vanilla':
      - conv_block_1
        - kernel_size (Choice: [3, 5, 7])
        - separable (Boolean)
        - max_pooling (Boolean)
        - num_blocks (Choice: [1, 2, 3])
        - num_layers (Choice: [1, 2])
        - filters_0_0 (Choice: [16, 32, 64, 128, 256, 512])
        - filters_0_1 (Choice: [16, 32, 64, 128, 256, 512])
        - dropout (Choice: [0.0, 0.25, 0.5])
        - filters_1_0 (Choice: [16, 32, 64, 128, 256, 512])
        - filters_1_1 (Choice: [16, 32, 64, 128, 256, 512])
    
    If block_type == 'resnet':
      - res_net_block_1 (Parent condition: image_block_1/block_type == 'resnet')
        - pretrained (Boolean)
        - version (Choice: ['resnet50', 'resnet101', 'resnet152', 'resnet50_v2', 'resnet101_v2', 'resnet152_v2'])
        - imagenet_size (Boolean)
    
    If block_type == 'xception':
      - xception_block_1
        - pretrained (Boolean)
        - imagenet_size (Boolean)
    
    If block_type == 'efficient':
      - efficient_net_block_1
        - pretrained (Boolean)
        - version (Choice: ['b0', 'b1', 'b2', 'b3', 'b4', 'b5', 'b6', 'b7'])
        - imagenet_size (Boolean)
- classification_head_1
    - spatial_reduction_type (Choice: ['flatten', 'global_max', 'global_avg'])
    - dropout (Choice: [0.0, 0.25, 0.5])
- optimizer (Choice: ['adam', 'sgd', 'adam_weight_decay'])
- learning_rate (Choice: [0.1, 0.01, 0.001, 0.0001, 2e-05, 1e-05])
    

\end{lstlisting}

\section{Search Space $\Phi _2$ for Hyperparameter Optimization}
\label{sec:appendix_b}
\begin{lstlisting}[style=mystyle]
- optimizer (Choice: ['adam', 'sgd', 'adam_weight_decay', 'rmsprop', 'adagrad', 'adadelta', 'nadam'])
- learning_rate (Choice: Real[0.1 ~ 1e-05])
- batch_size (Choice: [2, 4, 8, 16, 32])
- dropout (Choice: Real[0.0 ~ 0.5])

    

\end{lstlisting}
\setcounter{table}{0}

\section{Kfold Vs. Group Kfold Cross-Validation}
\label{sec:appendix_c}
\begin{table}[htpb]
\centering
\caption{Comparison of F1-score of 18 studied models on images from source 1 for two different cross-validation settings. CV indicates cross-validation.}
\label{tab:ttest}
\resizebox{0.7\textwidth}{!}{%
\begin{tabular}{@{}llll@{}}
\hline
Model Backbone                   & Variants          & Kfold CV    & Group Kfold CV \\ \hline
Xception                         &                   & 0.76 ± 0.08 & 0.75 ± 0.04    \\ \hline
MobileNet                        &                   & 0.80 ± 0.04 & 0.79 ± 0.03    \\ \hline
\multirow{6}{*}{ResNet}          & ResNet50          & 0.83 ± 0.03 & 0.81 ± 0.05    \\
                                 & ResNet50V2        & 0.79 ± 0.10 & 0.82 ± 0.02    \\
                                 & ResNet101         & 0.74 ± 0.08 & 0.73 ± 0.05    \\
                                 & ResNet101V2       & 0.81 ± 0.04 & 0.79 ± 0.06    \\
                                 & ResNet152         & 0.83 ± 0.02 & 0.80 ± 0.01    \\
                                 & ResNet152V2       & 0.83 ± 0.04 & 0.81 ± 0.09    \\ \hline
\multirow{2}{*}{Inception}       & InceptionV3       & 0.83 ± 0.03 & 0.82 ± 0.02    \\
                                 & InceptionResNetV2 & 0.76 ± 0.04 & 0.78 ± 0.01    \\ \hline
\multirow{3}{*}{DenseNet}        & DenseNet121       & 0.74 ± 0.11 & 0.75 ± 0.05    \\
                                 & DenseNet169       & 0.74 ± 0.12 & 0.73 ± 0.02    \\
                                 & DenseNet201       & 0.79 ± 0.04 & 0.80 ± 0.03    \\ \hline
\multirow{3}{*}{EfficientNet}    & EfficientNetB0    & 0.68 ± 0.09 & 0.70 ± 0.10    \\
                                 & EfficientNetB1    & 0.61 ± 0.11 & 0.62 ± 0.09    \\
                                 & EfficientNetB2    & 0.49 ± 0.14 & 0.52 ± 0.11    \\ \hline
\multirow{2}{*}{MoistNet (Ours)} & MoistNetLite      & 0.87 ± 0.05 & 0.86 ± 0.03    \\
                                 & MoistNetMax       & 0.91 ± 0.02 & 0.90 ± 0.04    \\ \hline
\end{tabular}%
}
\end{table}

\end{document}